\documentclass[journal]{IEEEtran}
\ifCLASSINFOpdf
\else
\fi

\usepackage{dblfloatfix}
\usepackage{amsmath}
\usepackage{amssymb}
\usepackage{color}
\usepackage{url}

\usepackage{graphicx}
\usepackage{subcaption}

\begin{document}

\title{Fast and Efficient Lenslet Image Compression}
\author{Hadi~Amirpour, ~\IEEEmembership{Student Member,~IEEE,}
Antonio~Pinheiro,~\IEEEmembership{Senior Member,~IEEE,}
Manuela~Pereira,
and~Mohammad~Ghanbari,~\IEEEmembership{Life Fellow,~IEEE}

\thanks{Hadi Amirpour, Antonio Pinheiro and Manuela Pereira are with Instituto de Telecomunica\c{c}\~{o}es and Universidade da Beira Interior, Covilh\~{a}, Portugal.
e-mail: (hadi.amirpourazarian@ubi.pt).}

\thanks{Mohammad Ghanbari is Professor at the School of Electrical and Computer Engineering, College of Engineering, University of Tehran, Tehran, Iran  (e mail:ghan@ut.ac.ir), as well as Emeritus Professor at the School of Computer Science and Electronic Engineering, University of Essex, Colchester, UK, CO4 3SQ, (email:ghan@essex.ac.uk).}
\thanks{The authors are very grateful to the Portuguese FCT-Funda\c c\~ao para a Ci\^encia e Tecnologia and co-funded by FEDER-PT2020, Portugal partnership agreement under the project PTDC/EEI-PRO/2849/ 2014 - POCI-01-0145-FEDER-016693, and under the project UID/EEA/50008/ 2013. }}

\markboth{}%
{Shell \MakeLowercase{\textit{et al.}}: Bare Demo of IEEEtran.cls for IEEE Journals}

\maketitle

\begin{abstract}
Light field imaging is characterized by capturing brightness, color, and directional information of light rays in a scene. This leads to image representations with huge amount of data that require efficient coding schemes. In this paper, lenslet images are rendered into sub-aperture images. These images are organized as a pseudo-sequence input for the HEVC video codec. To better exploit redundancy among the neighboring sub-aperture images and consequently decrease the distances between a sub-aperture image and its references used for prediction, sub-aperture images are divided into four smaller groups that are scanned in a serpentine order. The most central sub-aperture image, which has the highest similarity to all the other images, is used as the initial reference image for each of the four regions. Furthermore, a structure is defined that selects spatially adjacent sub-aperture images as prediction references with the highest similarity to the current image. 
In this way, encoding efficiency increases, and furthermore it leads to a higher similarity among the co-located Coding Three Units (CTUs). The similarities among the co-located CTUs are exploited to predict Coding Unit depths.    
Moreover, independent encoding of each group division enables parallel processing, that along with the proposed coding unit depth prediction decrease the encoding execution time by almost $80\%$ on average. Simulation results show that Rate-Distortion performance of the proposed method has higher compression gain than the other state-of-the-art lenslet compression methods with lower computational complexity.
\end{abstract}


\IEEEpeerreviewmaketitle

\section{Introduction}
\label{sec:introduction}
\IEEEPARstart{R}{ecent} 
photography technologies can produce richer spatial information in order to increase the 3D world representation of images. With the introduction of commercial light field cameras into the consumer market such as Lytro\footnote{https://www.lytro.com/} and Raytrix\footnote{http://www.raytrix.de/} light fields have recently gained a lot of attentions as an emerging technology. Light field cameras capture the intensity information as well as directional information, enabling post-capture processing like re-focusing, new views synthesizing, holographic reconstruction, and depth estimation~\cite{Overview}. 

\noindent From a theoretical viewpoint, light field is defined by the plenoptic function ($P$) \cite{plenoptic} as the intensity of light rays in space passing through every point ($V_x, V_y, V_z$) at every direction ($\theta, \phi$), for every wavelength $\lambda$, at any time $t$:
\begin{equation}
P=P(\theta, \phi, \lambda, t, V_x, V_y, V_z)
\end{equation} 
\noindent However, recording and manipulating such high dimensional information is difficult and some simplifications are commonly used. The first simplification considers the function to be time-invariant and monochromic, removing $t$ and $\lambda$ dimensions, respectively. The second simplification assumes the space is free of occlusion resulting in a simplified function of four dimensions, 
\begin{equation}
P=P(u, v, r, s )
\label{Psimpl}
\end{equation} 

\begin{figure}[!b]
\center
\begin{minipage}[b]{\linewidth}
\centering
\includegraphics[width=0.6\textwidth]{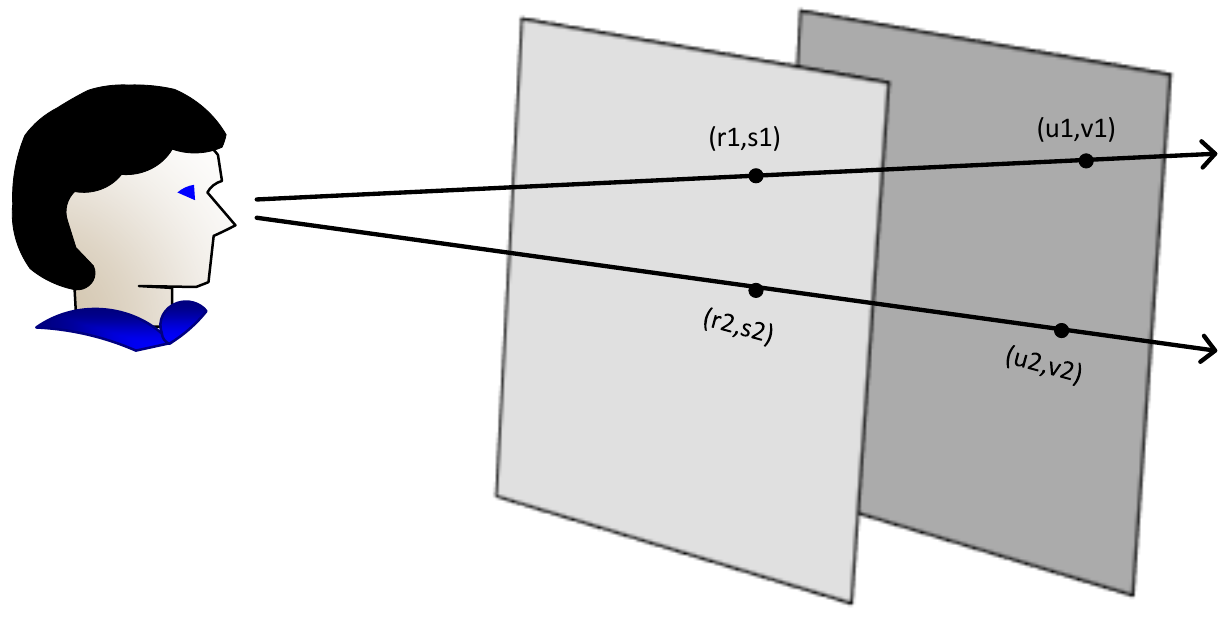}
\end{minipage}
\caption{Parametrization of rays by using two parallel $uv$ and $rs$ planes leading to the so-called light field. }
\label{fig:3}
\end{figure}

\noindent 
Equation \ref{Psimpl} defines that each reflected ray by the object passes through a $uv$ plane and intersects an $rs$ plane (Fig.~\ref{fig:3}). The $rs$ plane can be considered as a set of cameras which collect the rays leaving from the $uv$ plane. Therefore, each point on the $rs$ plane collects a set of rays that constitute a viewpoint. 

Multi-camera arrays and lenslet imaging are the main light field imaging acquisitions \cite{1673328,Overview25,Wetzstein}. Lenslet light field cameras have advantages like in-camera light field imaging, portability, and cost-effectiveness especially in high-resolution imaging, becoming widely used for the light field image acquisitions. They have a micro-lens array between the camera's main lens and the photosensor obtaining angular information of scene together with the spatial information.
The Lytro Illum\footnote{https://www.lytro.com/} captures 2D lenslet raw images with 7728 $\times$ 5368 pixels resolution. 
This rich information results in a huge amount of data that challenges the practical use of these images, making the design of a highly efficient and low complexity coding scheme necessary. In this paper, a highly efficient compression method for the lenslet images is introduced based on the state-of-the-art High Efficiency Video Coding (HEVC) codec \cite{HEVC}. The major contributions include the definition of a new scanning order that enables improved compression efficiency as well as parallel processing, and a new referencing structure that provides improved compression and computational efficiency. To reduce time-complexity a CTU depth prediction method is proposed. 
The remaining part of the paper is organized as follow: Section~\ref{sec:R_W} presents recent approaches to light field compression and Section~\ref{sec:P_M} introduces the proposed approach. Experimental results are provided in Section~\ref{sec:S_R}. Section~\ref{sec:conclusion} concludes the paper.

\section{Related works}
\label{sec:R_W}

Various approaches for compression of lenslet images have been proposed. These approaches are categorized into two main classes: direct and indirect. In the direct methods, the compression is applied directly to the raw lenslet image. In the indirect methods, the compression is applied to the sub-aperture images rendered from the raw lenslet image. 
In the direct approaches, in addition to the use of conventional still image coding methods, like JPEG, JPEG2000 or HEVC intra coding, several proposals directly exploit the existing spatial redundancy among the microlens images within a raw lenslet image~\cite{Conti2012, Conti2016,Conti2016Bi,Conti2016LLE,Conti2014}. 

The disparity-compensated prediction method in~\cite{Disparity} tries to use the existing spatial redundancy among the microlenses in a lenslet image. Higher order intra block prediction method in~\cite{HOP} adds Higher Order Prediction (HOP) to HEVC intra prediction modes through a geometric transformation between the current block and the reference region.

\begin{figure}[!b]
\centering
\includegraphics[width=0.5\textwidth]{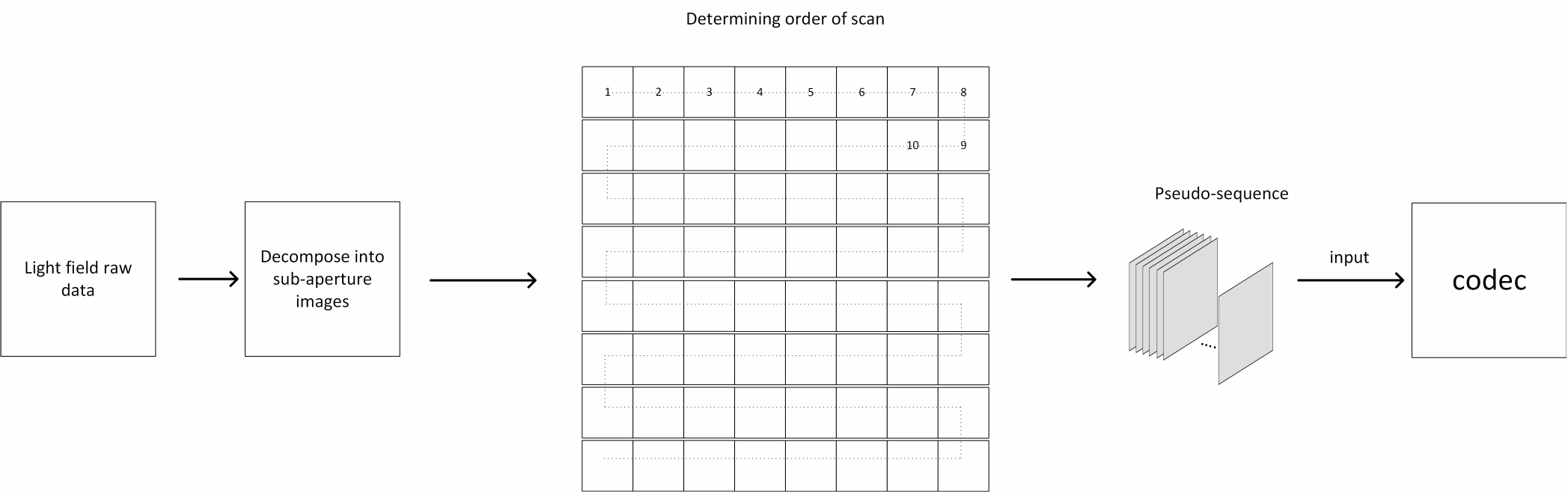}
\caption{Converting a raw lenslet image to a sub-aperture pseudo-sequence used as input of a video codec. }
\label{fig:4}
\end{figure}

The pseudo sequence-based approaches are the most popular indirect methods for light field image compression. 
Perra \textit{et al.}~\cite{Tiling} partition the raw light field image into equal size tiles and then order them as pseudo-temporal sequences to be inter picture coded by HEVC encoder.
These methods define a sequence with the different views of the light field image that is compressed as a video sequence.
In case of lenslet images, they are initially rendered into sub-aperture images (as shown in Fig.~\ref{fig:4}). A video codec (like for instance HEVC) encodes this sequence of sub-aperture images with a predefined scan order.

To define a higher redundancy among sub-aperture pseudo-sequences and increase the compression efficiency several papers have reported the on-going research of the sub-aperture image sequence strategy. Raster and spiral scans have been compared in~\cite{Raster}. Raster, serpentine, zigzag and circular scans have been compared in~\cite{Circular}. In~\cite{Snake}, serpentine and spiral scans using HEVC with different reference structures are compared. A 2-D hierarchical coding structure for the light field image compression is used in~\cite{2D}.

Considering $13\times13$ out of $15\times15$ sub-aperture images, the four border sub-aperture images of top-left, top-right, bottom-left and bottom-right are typically rendered as dark and represent less information. They also have the lowest similarity with their adjacent sub-aperture images. 
In raster or serpentine scan orders, the top-left sub-aperture image is selected as the first frame and is intra coded. However, this sub-aperture image is the sub-aperture image that has less information and less similarity with the other sub-aperture images. Moreover, using it as the reference image for inter picture coding of the other sub-aperture images, will reduce their reconstruction quality. 
However, suppressing these sub-aperture images in the coding process will lead to loss of information. The scans like spiral that start from the most central sub-aperture image, unlike raster and serpentine scans, starts ordering the sub-aperture images from the central sub-aperture image. This sub-aperture image is intra coded and does not lead to the above mentioned problems.

\begin{figure}[!b]
\begin{minipage}{1.0\linewidth}
\centering
\centerline{\includegraphics[width=0.65\textwidth]{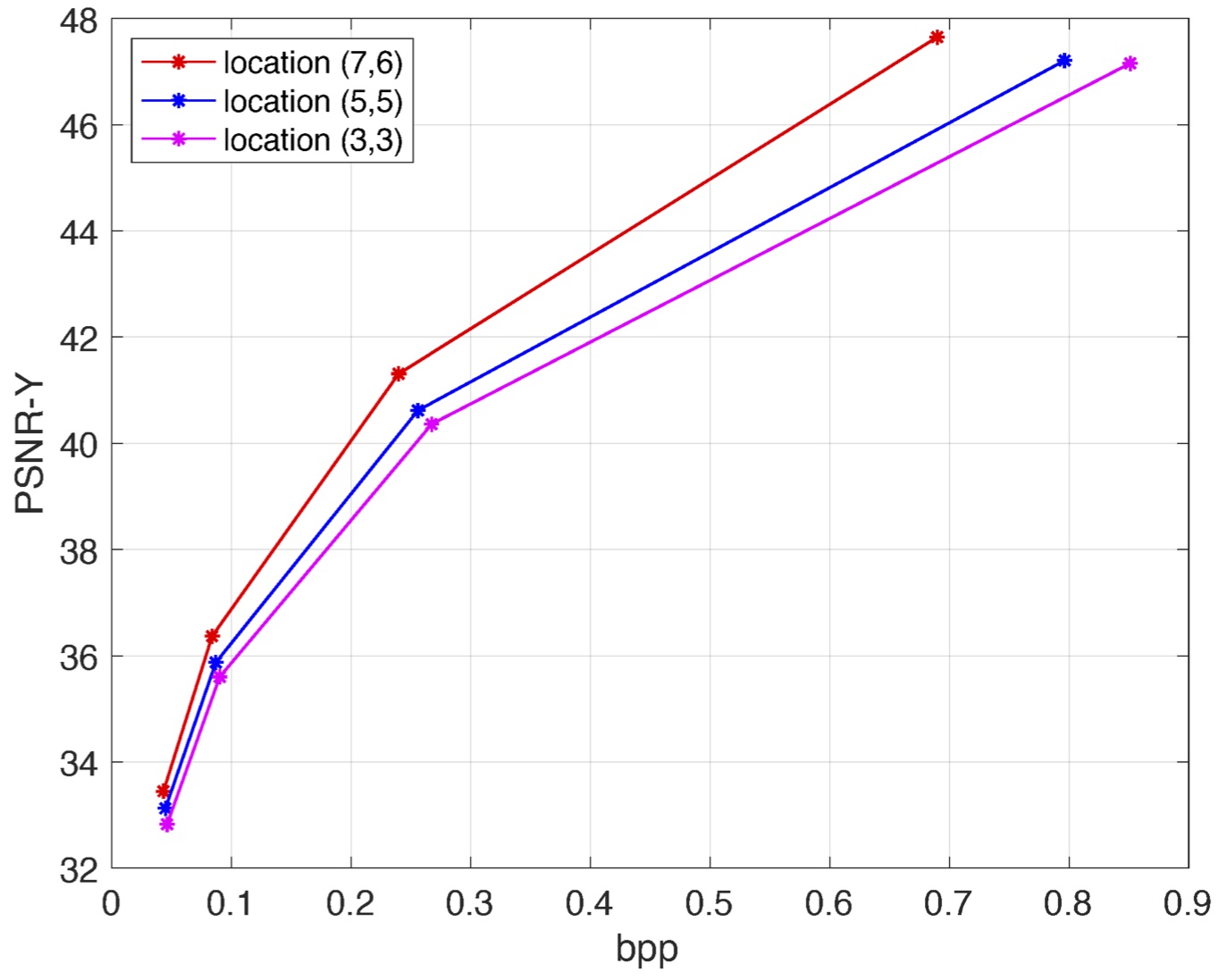}}
\end{minipage}
\caption{Importance of spatial distance in coding efficiency.}
\label{fig:dis}
\end{figure}

The similarity between a sub-aperture image and its references is also an important parameter for the compression efficiency when inter-frame coding of a video codec is used. 
This similarity decreases as the distance between sub-aperture images increases. Fig.~\ref{fig:dis} shows an example where three sub-aperture images located at different distances are inter-frame encoded using the same most central sub-aperture image, located in $(7,7)$, as the reference. It can be observed that the closer is the sub-aperture image of the reference the higher is the compression performance. 

In this paper, a new methodology for lenslet image compression is proposed, which reduces the distance between images and their references and avoids using dark border images as references.

\begin{figure}[!b]
\centering
\begin{subfigure}[b]{0.2\textwidth}
\includegraphics[width=\textwidth]{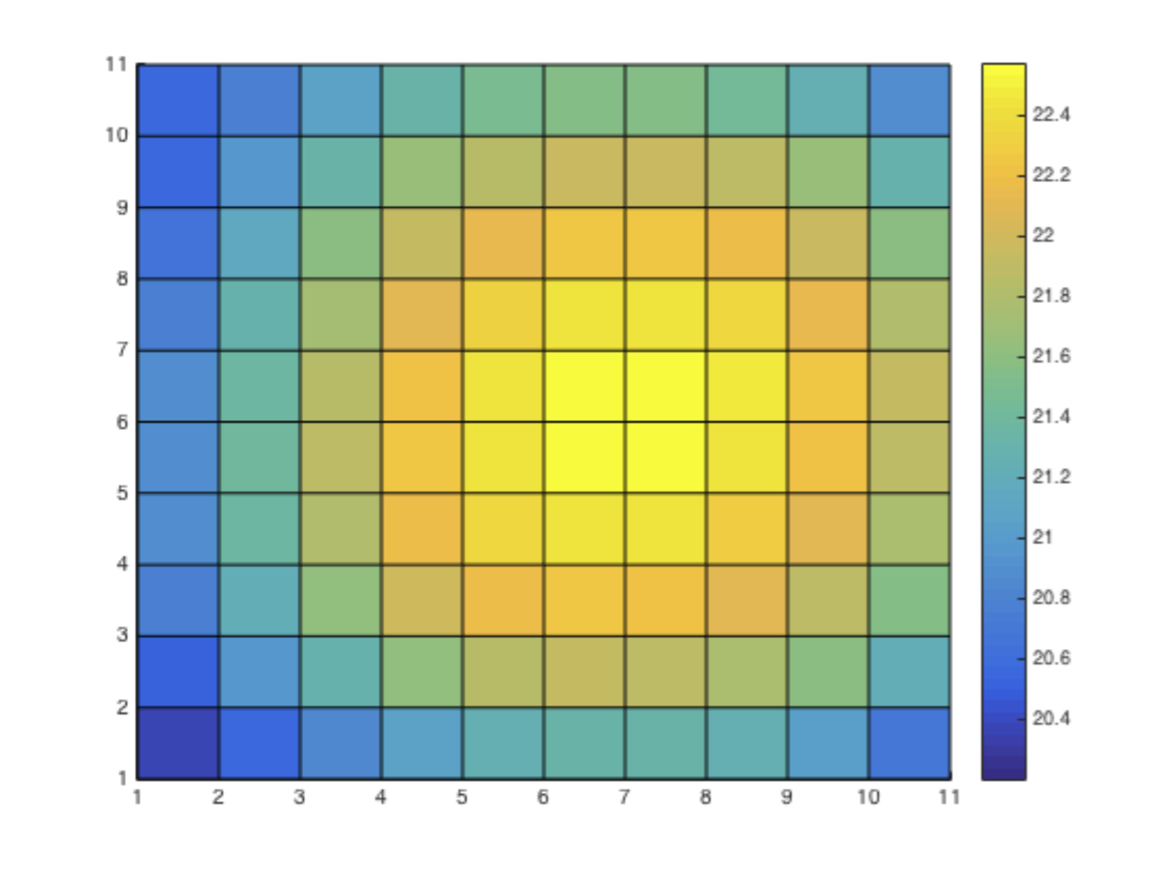}
\caption{Bikes}
\label{fig:gull}
\end{subfigure}
~ 
\begin{subfigure}[b]{0.2\textwidth}
\includegraphics[width=\textwidth]{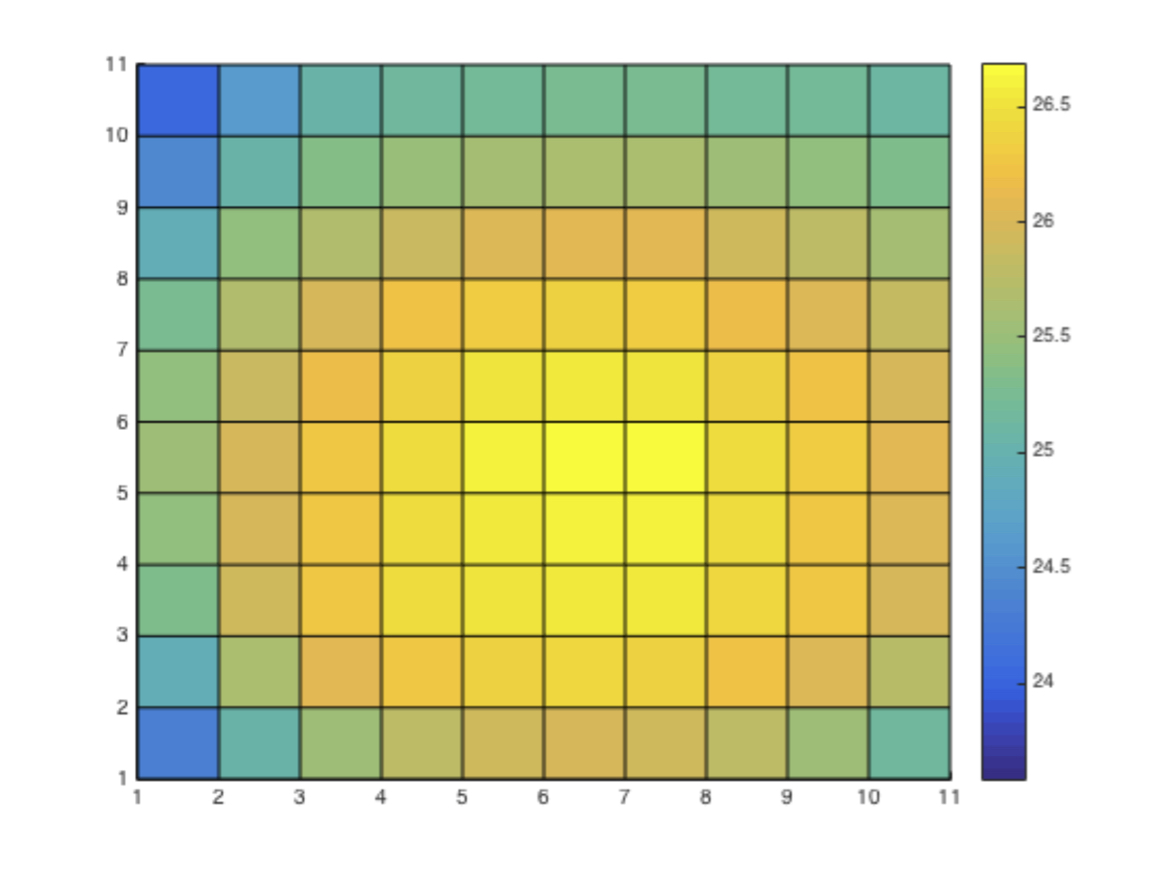}
\caption{Friends-1}
\label{fig:tiger}
\end{subfigure}
~ 
\begin{subfigure}[b]{0.2\textwidth}
\includegraphics[width=\textwidth]{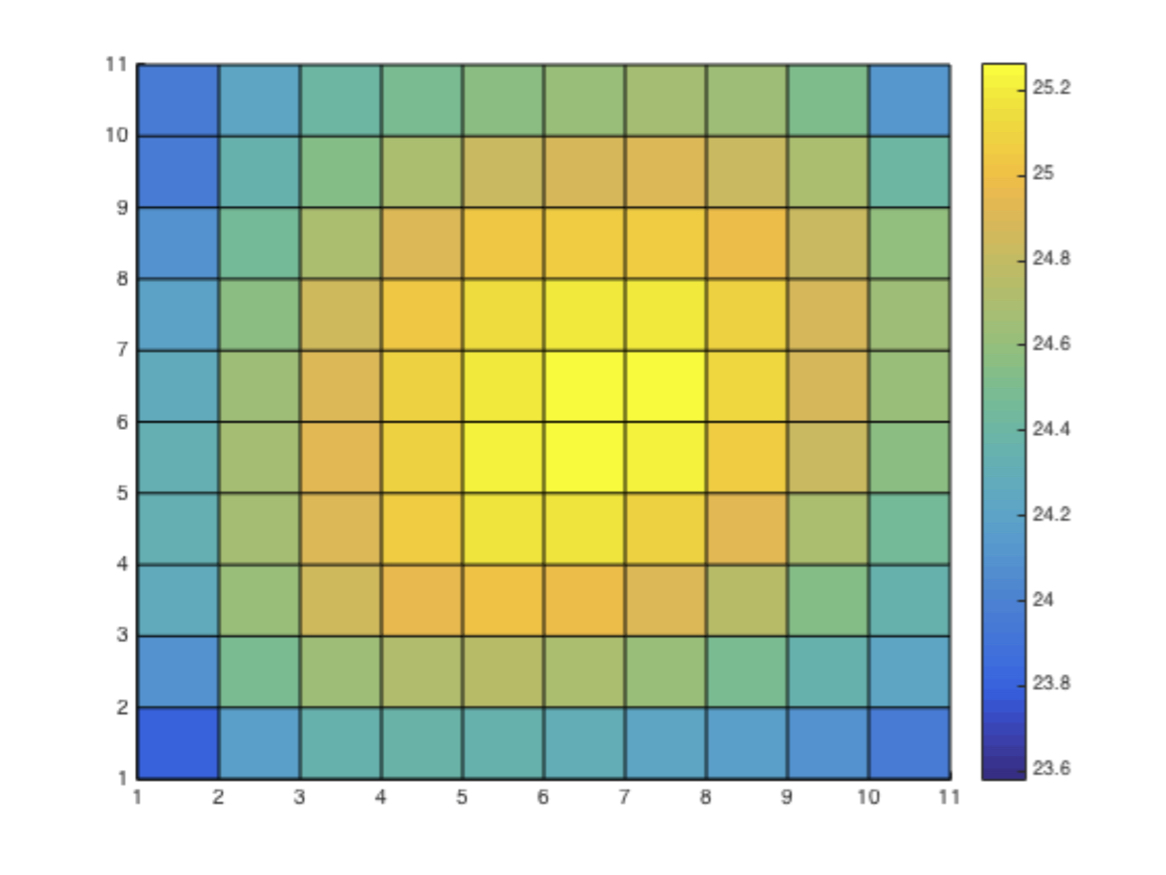}
\caption{Stone\_Pillars\_Outside}
\label{fig:mouse}
\end{subfigure}
~ 
\begin{subfigure}[b]{0.2\textwidth}
\includegraphics[width=\textwidth]{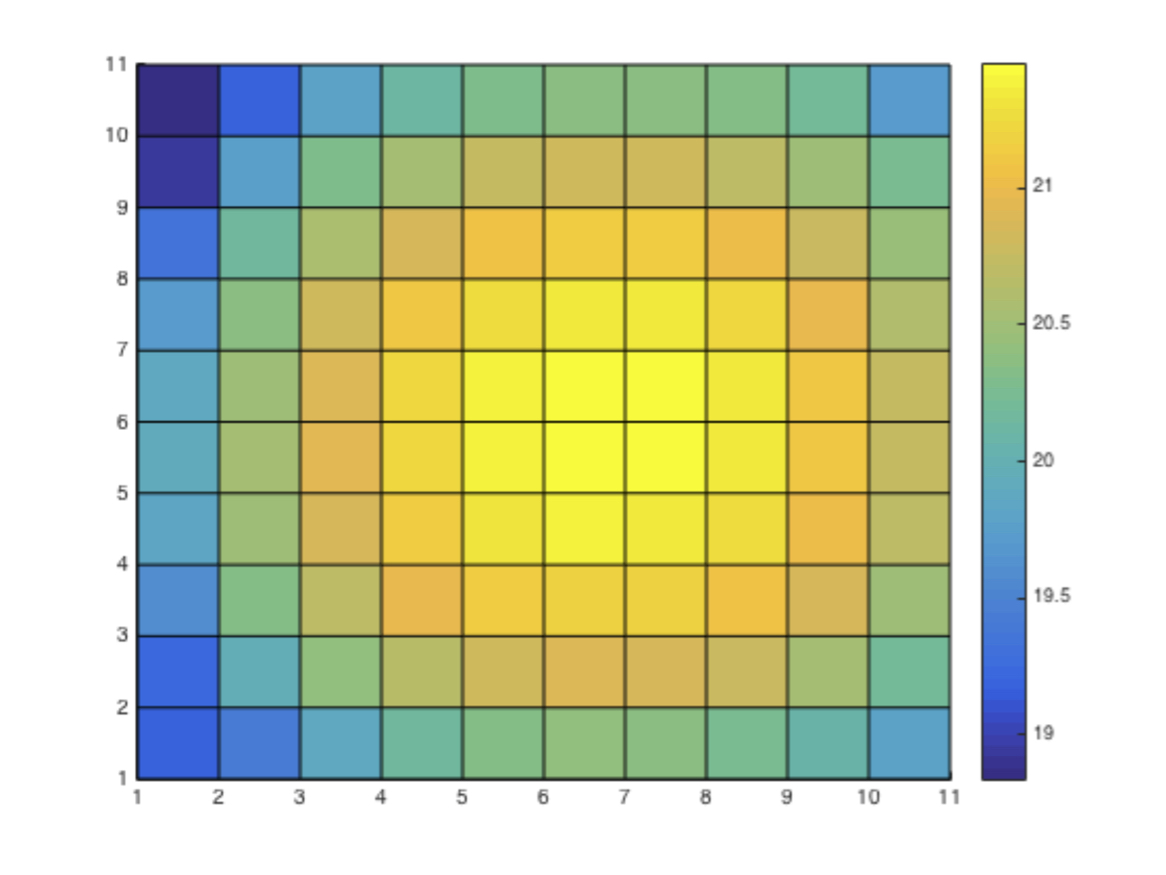}
\caption{Fountain\&Vincent\_2}
\label{fig:mouse}
\end{subfigure}
\caption{Representation of the average PSNR value between each sub-aperture image and all the others.}
\label{fig:10}
\end{figure}

\section{Proposed method}
\label{sec:P_M}

\subsection{Coding order}
\label{sec:CO}

The first image to be encoded is more important than the other images. Due to its error propagation, its reconstruction quality will affect reconstruction quality of the remaining images. Furthermore, its similarity with the other images will have a significant impact on their reconstruction quality. To compute the similarity between each sub-aperture image with the other sub-aperture images, PSNR between each sub-aperture image and all the others is computed. 
Then, the average of the PSNR values for each sub-aperture image is computed. 
These average PSNR values are represented in Fig.~\ref{fig:10} for four lenslet images (Yellow shows largest average PSNR, while blue shows smallest average PSNR). This representation supports the fact that the central sub-aperture images typically have a larger average similarity to the other sub-aperture images. The average value decreases, as sub-aperture images become further away from the central sub-aperture images.

Therefore, unlike the conventional scan methods, as the serpentine and raster searches that start from upper-left sub-aperture (which is typically a dark image for the lenslet images, with small similarity with the other sub-aperture images), we propose to encode the most central sub-aperture image in the intra coding (as the first reference image). This sub-aperture image tends to be more similar to the other sub-aperture images leading to a closer prediction of the remaining sub-aperture images.

\begin{figure}[!b]
\begin{minipage}{1.0\linewidth}
\centering
\centerline{\includegraphics[width=0.65\textwidth]{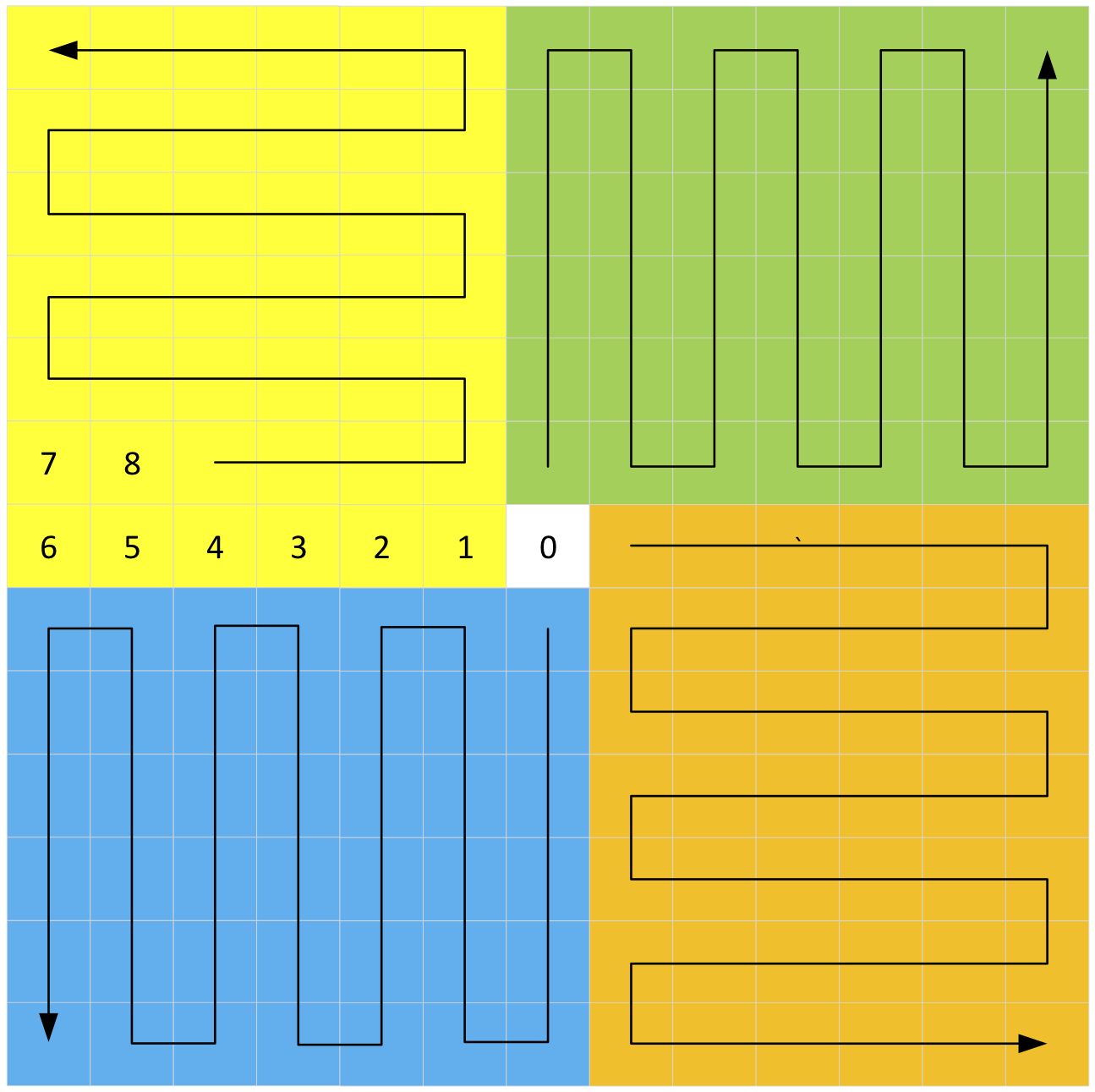}}
\end{minipage}
\caption{Proposed scan order.}
\label{fig:8}
\end{figure}

In the conventional scan orders that use the group of pictures (GOP) structure of HEVC, such as low-delay predictive coding configuration, the distance of a current sub-aperture image with its references is large in most of the cases. These larger distances result in less efficient compression when compared with the use of spatially closer references, as they present higher similarity that can be efficiently used in the prediction mechanism. To reduce the distance between the current frame and its reference frames, sub-aperture images are divided into four smaller groups and each group is encoded independently. First, 13$\times$13 middle sub-aperture images out of 15$\times$15 whole sub-aperture images are selected. For each lenslet image four 6$\times$7 divisions of sub-aperture images are defined together with the central sub-aperture image, resulting in the whole set of 13$\times$13 images. This structure is shown in Fig.~\ref{fig:8}. The sub-aperture images are scanned horizontally in serpentine order in two of the divisions, and are scanned vertically in serpentine order in the other two divisions. Hence, unlike conventional scanning methods which scan sub-aperture images in a 13$\times$13 region, the proposed method scans them into four independent divisions of 6$\times$7 sub-aperture images. This way, average distance between sub-aperture images and their references will decrease significantly.  The smaller distance for the proposed scan order compared with the conventional scan orders results in a higher similarity between the current sub-aperture image and the used references.

Moreover, the previously mentioned low similarity problem of the border sub-aperture images with the others is also avoided. The developed method does not use border sub-aperture images as references, in contrast to the conventional HEVC scanning orders, which adversely affect the compression efficiency. 
For instance, the raster method uses the upper-left sub-aperture as the first image of the sequence. Hence, it is encoded with an HEVC codec in intra mode. The following sub-aperture images in the sequence will have smaller similarities to the first scanned image and hence the compression efficiency of the inter coding mode in the following frames will be reduced. However, in the proposed method, the corner sub-aperture images are the last sub-apertures to be coded in each region, and hence their negative impacts on compression efficiency are minimal, as it can be seen in Fig.~\ref{fig:8}. 
Furthermore, since the most central sub-aperture image is intra coded, they have better quality. Using it as the first reference image will lead to a better reconstruction quality, too. Using this structure, the intra coded sub-aperture image is coded once, but it is used four times as the first reference due to the independent coding of each quadrant. 
\begin{figure}[!t]
\begin{minipage}{1.0\linewidth}
\centering
\centerline{\includegraphics[width=8.5cm,height=5cm]{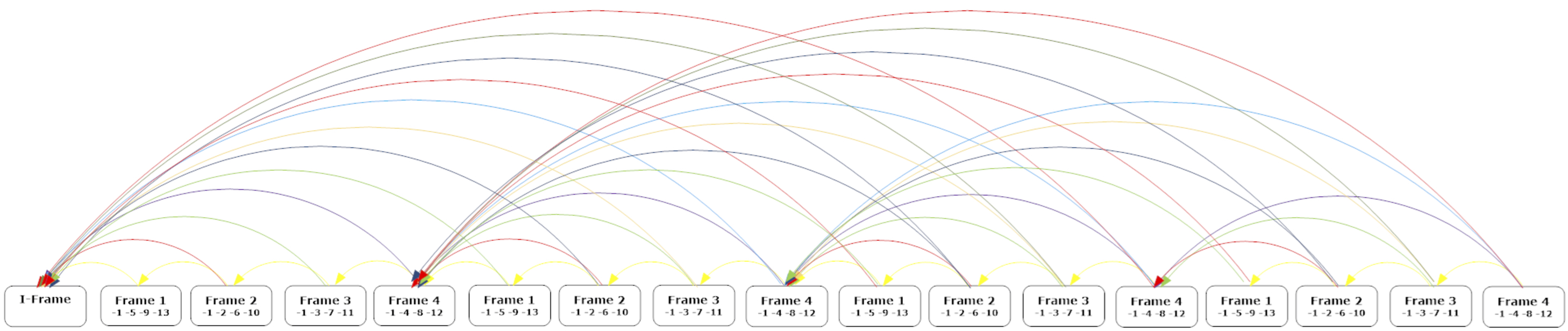}}
\end{minipage}
\caption{Low-delay prediction.}
\label{fig:6}
\end{figure}

\subsection{Reference structure}
\label{sec:rs}

\begin{figure}[!b]
\begin{minipage}{1.0\linewidth}
\centering
\centerline{\includegraphics[width=0.5\textwidth]{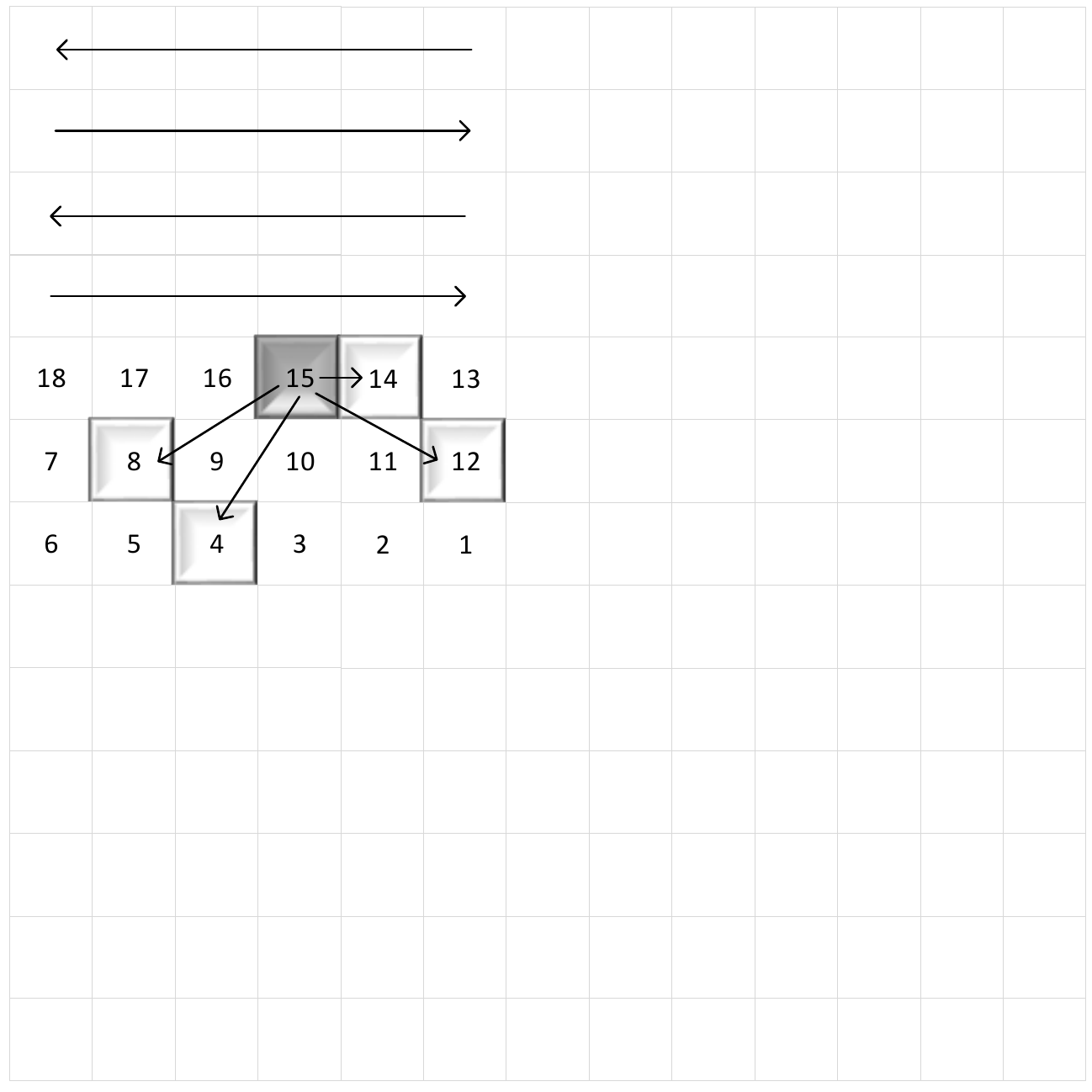}}
\end{minipage}
\caption{$15^{\mbox{th}}$ sub-aperture image's references using low-delay prediction}
\label{fig:11}
\end{figure}

HEVC low-delay prediction uses a GOP structure with four reference frames for each inter coded frame. This structure is illustrated in Fig.~\ref{fig:6}. Conventional lenslet compression methods use this structure to encode the images. This reference structure has been designed to encode real video sequences and uses temporally distant frames to determine the references. However, distances in sub-aperture images are spatially defined, not temporally. Hence, the GOP structure designed for video sequences is not suitable for sub-aperture pseudo-sequence lenslet image coding. 
Fig.~\ref{fig:11} shows an example where sub-aperture images 14, 12, 8, and 4 are the references for sub-aperture image 15 considering the low-delay prediction structure illustrated in Fig.~\ref{fig:6}. In this example, only one of the reference images (14) is an adjacent sub-aperture image. 

\begin{figure}[!t]
\centering

\begin{subfigure}[b]{1\linewidth}
\center
\includegraphics[width=0.5\textwidth]{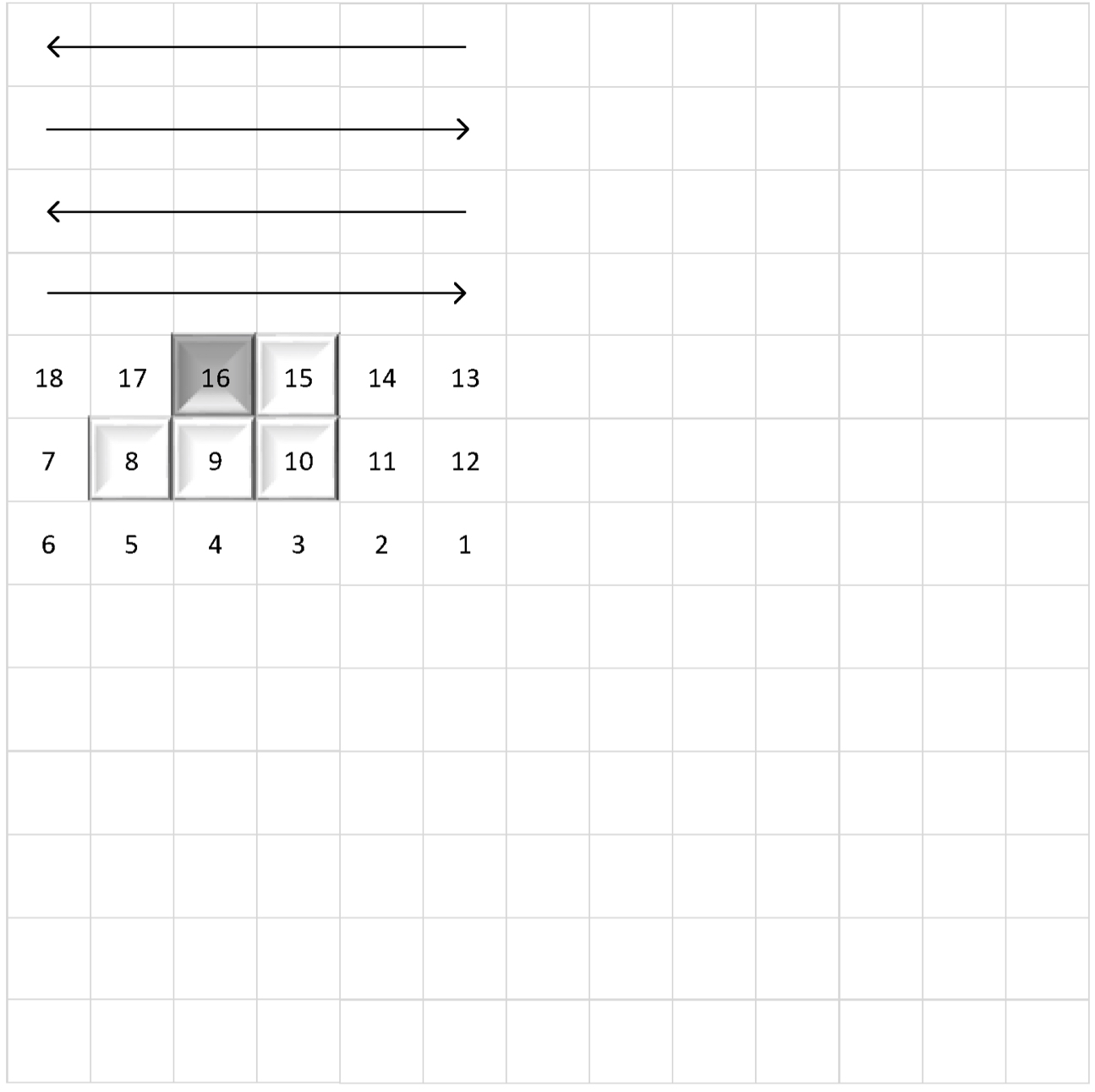}
\caption{}
\label{fig:10-4}
\end{subfigure}

\caption{An example of proposed referencing for one GOP.}
\label{fig:12}
\end{figure}

To better exploit redundancy among the sub-aperture images a new reference structure is proposed in this paper. 
It uses distances between sub-aperture images to select sub-aperture image references localized in adjacent positions, instead of using the Picture Order Count (POC) of HEVC. 
This strategy improves the coding efficiency, due to a better exploitation of the similarity between sub-aperture images. 
Fig.~\ref{fig:12} shows examples of the proposed referencing method. This model is generalized for all the sub-aperture images in the different quadrants. Hence, the average distance between the reference sub-aperture images and the uncompressed one will decrease to lower values.
\begin{figure}[!b]
\begin{minipage}[!b]{1.0\linewidth}
\centering
\centerline{\includegraphics[width=\textwidth]{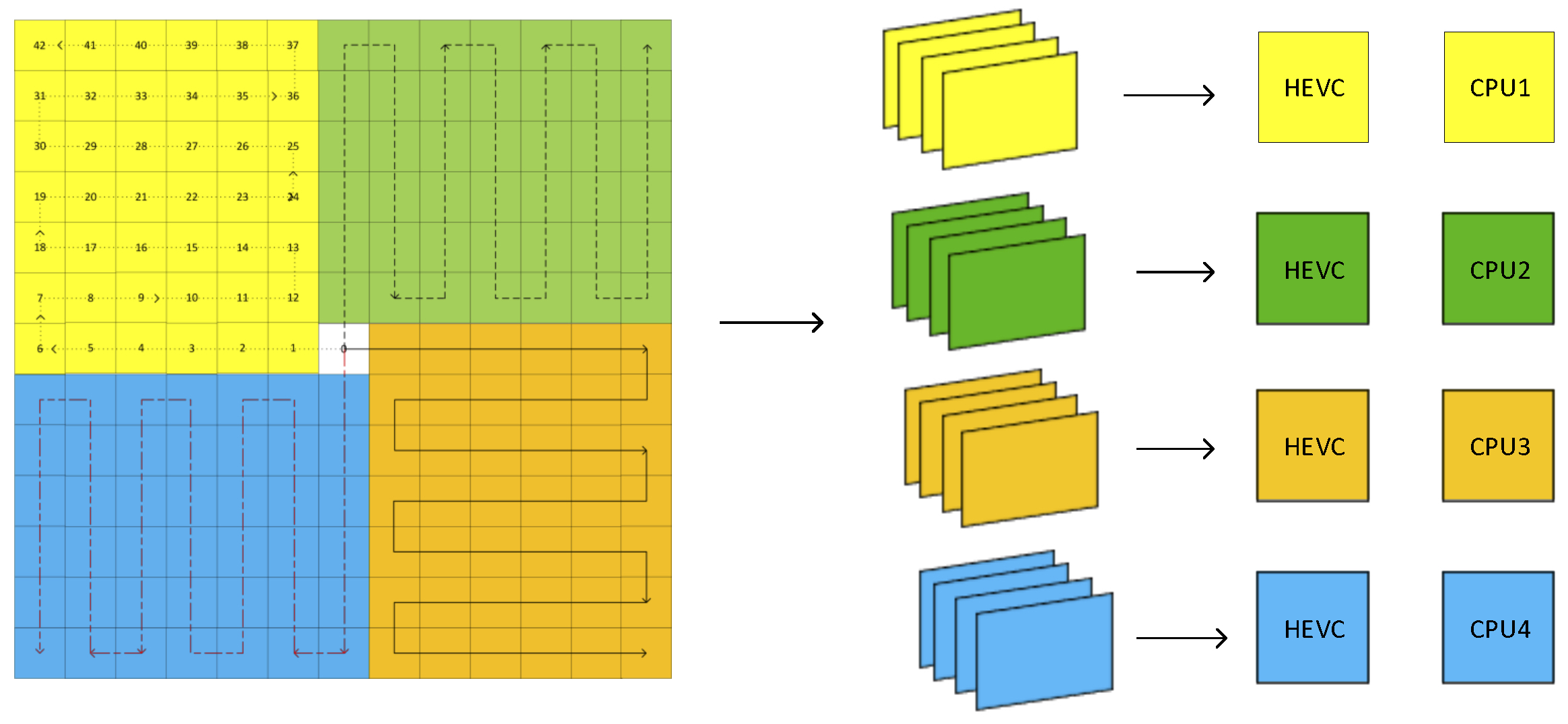}}
\caption{Workflow for the parallel processing.}\medskip
\label{fig:par}
\end{minipage}
\end{figure}

\subsection{Parallel processing}
\label{sec:par}

Besides coding efficiency, encoding/decoding time complexity, as well as memory footprint are major challenges in light field compression. Huge number of sub-aperture images in addition to their higher resolutions makes their encoding an extremely time-consuming process. Encoding of each sub-aperture image depends on previously encoded images, therefore, they will be encoded in serial chain processes with conventional pseudo-sequence based methods. However, the proposed method encodes each region independently. Independent encoding of regions enables using parallel processing. Fig.~\ref{fig:par} shows the workflow for parallel processing. Sub-apertures in each colored group make a pseudo-video and four different processors are used to encode each pseudo-video.
In addition, the proposed method has faster access to individual view than the other pseudo-sequence methods due to independent encoding/decoding of each region. Access to a sub-aperture image with the conventional scanning orders~\cite{ Tiling, Raster, Circular} needs decoding and loading 168 views into memory, while for the proposed method only 42 views are required. Therefore, decoding one individual view will require approximately four times less memory than the conventional pseudo-sequence methods.

\subsection{CTU partitioning }
\label{sec:par}

\begin{figure}[!b]
    \centering
        \includegraphics[width=0.5\textwidth,height=0.2\textwidth]{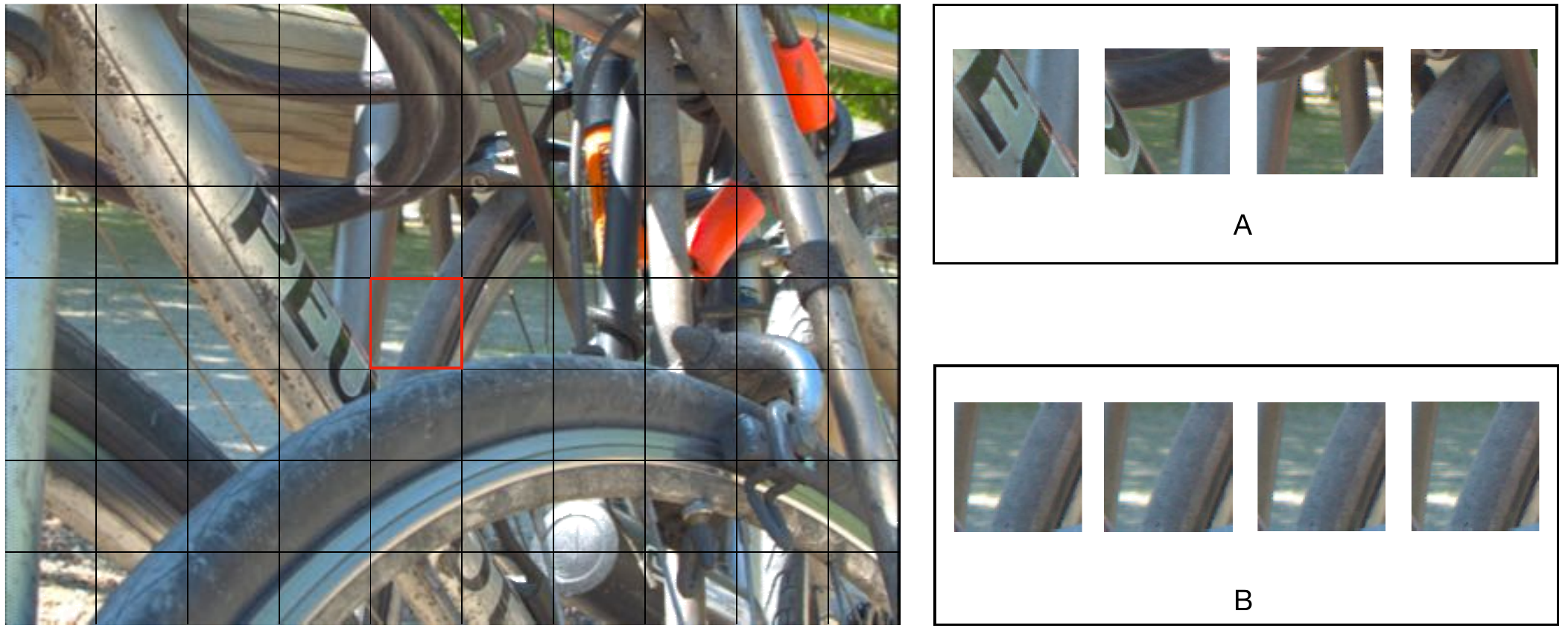}
    \caption{Block partitioning of one lenslet image. Current CTU represented with red borders and its four spatially neighbouring CTUs are shown in (A). Its four co-located CTUs in the spatially closer sub-aperture images are shown in (B).}
    \label{fig:partition}   
\end{figure}

The higher similarity between spatially closer sub-aperture images selected as reference images can be exploited to predict depth partitioning of the encoding sub-aperture image.
Fast coding unit (CU) depth decision algorithms mainly use depth information of the neighbouring CTUs to predict the depth range of the current CTU. In Fig.~\ref{fig:partition}, the most central sub-aperture image of a lenslet image with resolution $626\times434$ pixels has been divided into CTUs. Fig.~\ref{fig:partition}(A) shows a CTU and its four spatially neighbouring CTUs. It is easily seen that CTUs have different textures and may be encoded at different depths. Fig.~\ref{fig:partition}(B) shows the same CTU with four co-located CTUs in four spatially closer reference images. It can be observed that co-located CTUs especially those which are selected from the spatially closer images have a higher similarity. In addition, CTUs located in the last row and column are not of the same size of the other CTUs and cannot use CTU information of some of the spatially neighbouring CTUs. Hence, these CTUs are used to predict depth of the current CTU. 

Conventional methods in video coding, typically use the maximum depth of a CTU as its depth representation. They then use it to predict depth of other CTUs~\cite{ctu_tm,ctu_ip,ctu_tii}. However, each CTU may have different regions ranging from homogeneous to complex textures. Therefore, determining the depth range of a CTU may not be precise for all the CUs. 

\begin{figure}[!b]
    \centering
        \includegraphics[width=0.5\textwidth]{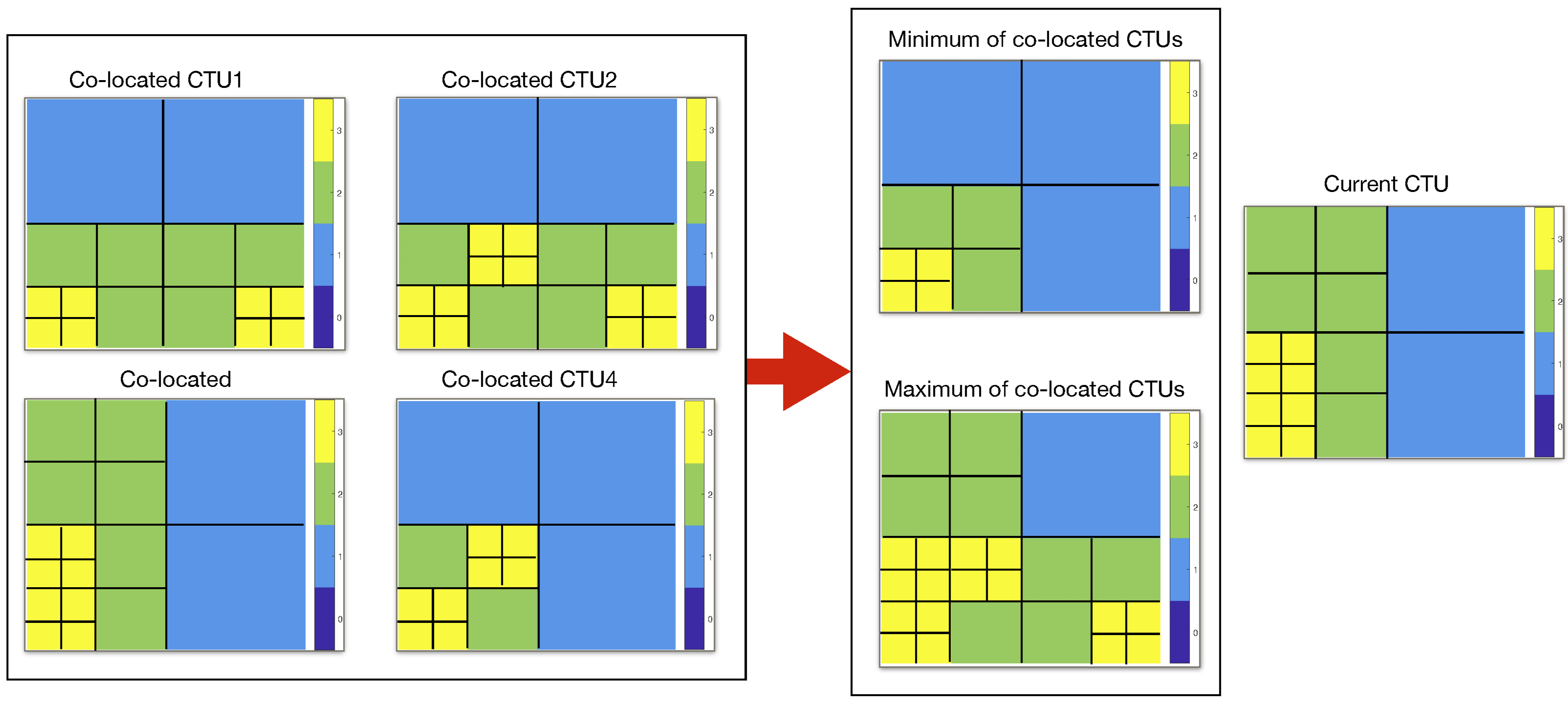}
    \caption{Minimum and maximum depths of co-located CTUs used to limit depth of the current CTU.}
    \label{fig:depth}
\end{figure}

In this study, a different depth is predicted for each $8\times8$ CU. For instance, depth maps for four co-located CTUs located in the four spatially closer sub-aperture images are shown in Fig.~\ref{fig:depth}. To predict the CU level depth map for the current CTU, the minimum and maximum depths of co-located CTUs are used to limit the depth of CUs in the current CTU. Fig.~\ref{fig:depth} shows an example for the proposed method. The so-called current CTU, and its four co-located CTUs in four spatially closer sub-aperture images have been encoded using the HEVC reference software. It is obviously seen that the depth map for CUs in the current CTU is limited between the minimum and maximum depths of the co-located CTUs. Hence, depths that are smaller than $depth_{min}$ and larger than $depth_{max}$ for each CU will be ignored. 
The CTU depth range for the example of Fig.~\ref{fig:depth} will be [1, 3]. Hence, depth 0 zero will be skipped. However, using the proposed method different regions of a CTU will have different depth ranges resulting in a reduction on the overall search. Considering Fig.~\ref{fig:depth} example, depth for all the co-located CTUs in the upper-right region is 1 and depth range will be [1 1] for CUs in this region. Depths 0, 2, and 3 will be ignored for the CUs in that region.
\begin{figure}[!b]
    \centering
        \includegraphics[width=0.3\textwidth]{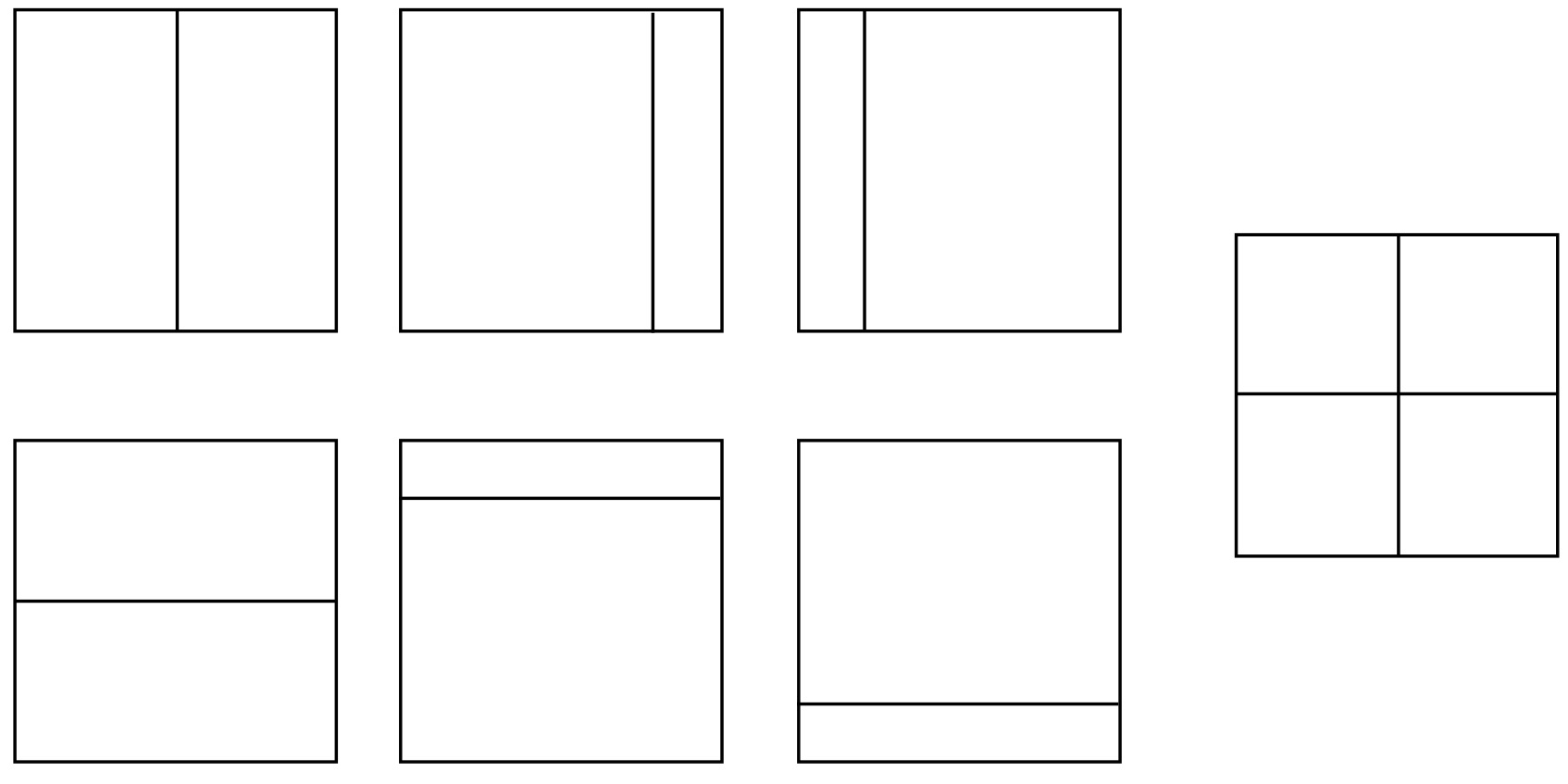}
    \caption{non-$2N\times 2N$ PUs.}
    \label{fig:non2n}
\end{figure}
\subsubsection{Depth for non-$2N\times 2N$ Partition Unit sizes}
One CU ($2N\times 2N$) in each depth can have different Partition Unit (PU) sizes including $2N\times 2N$, $N\times 2N$, $nR\times 2N$, $nL\times 2N$, $2N\times N$, $2N\times nT$, $2N\times nD$ and $N\times N$. 
The selected depth ($d$) is considered whatever is the PU size.
However, non-$2N\times 2N$ cases, as shown in Fig.~\ref{fig:non2n}, were divided into smaller parts. In the developed method, depth ($d$) for the CUs that have non-$2N\times 2N$ PU size will be incremented when they are used to predict the current CTU. So, depth of CUs is computed as follows:

\begin{equation*}
d = \begin{cases}
d & \text{PU size is $2N\times2N$}\\
d+1 &\text{PU size is not $2N\times2N$}\
\end{cases}
\end{equation*}

\subsubsection{PU Histogram} 
\begin{figure}[!h]
    \centering
        \includegraphics[width=0.3\textwidth]{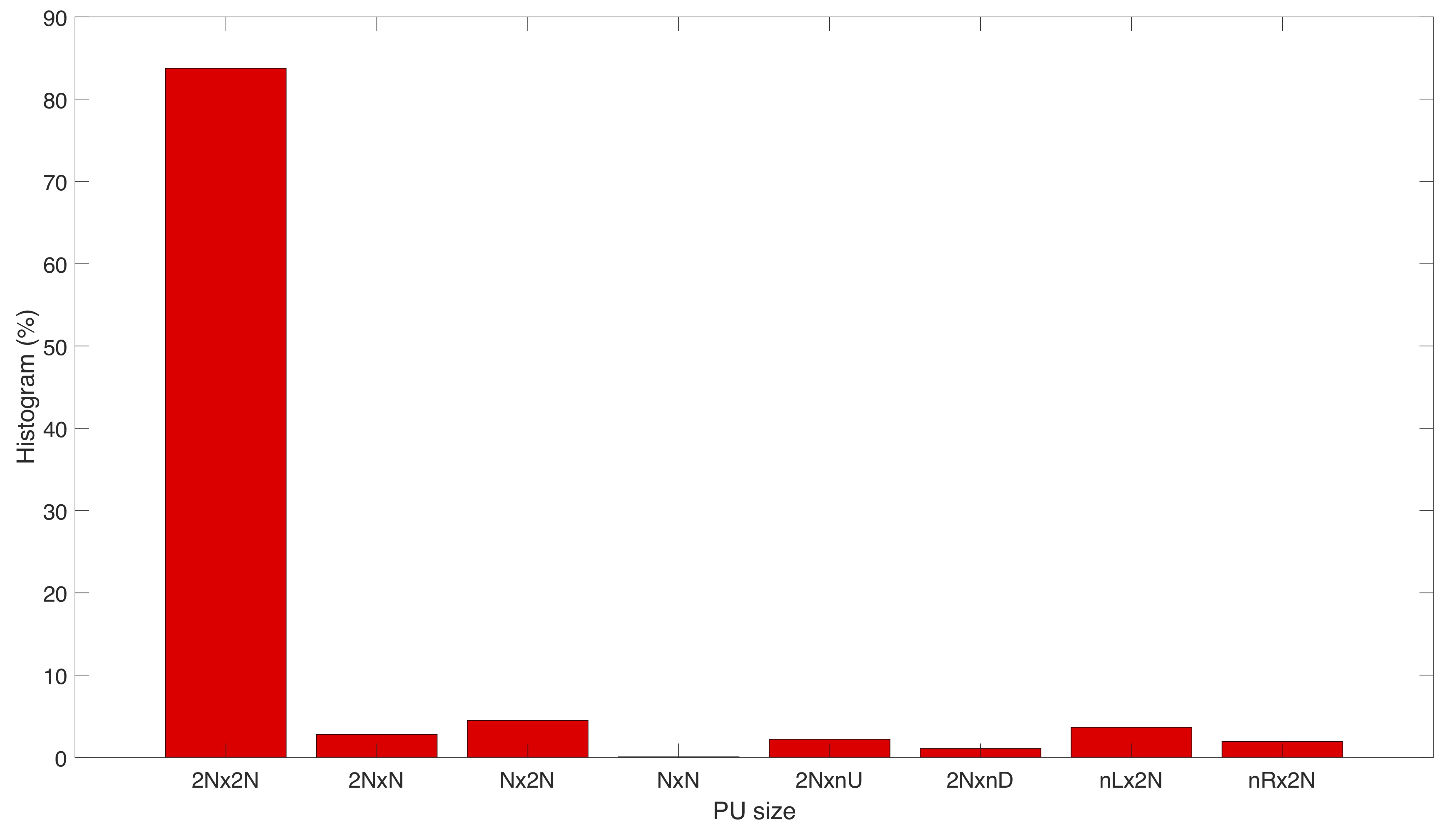}
    \caption{Histogram of PU sizes.}
    \label{fig:pu}
\end{figure}

\begin{figure}[!h]
    \centering
        \includegraphics[width=0.4\textwidth]{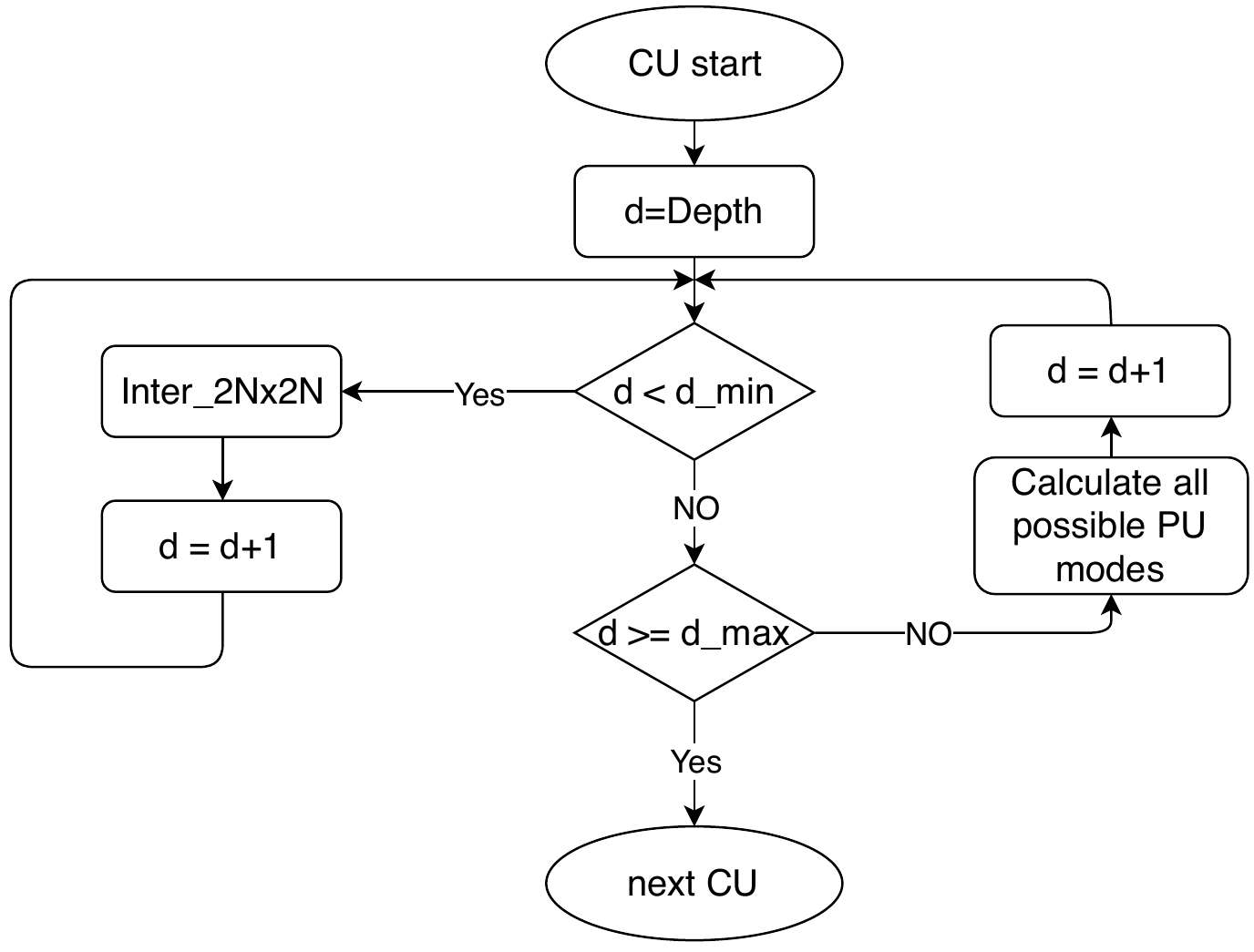}
    \caption{Flowchart of proposed CU depth decision.}
    \label{fig:flowchart}
\end{figure}

As discussed previously, the proposed method skips all PU modes with depths smaller than $depth_{min}$. However, for $d<depth_{min}$ only the $2N\times 2N$ PU mode is searched to increase the accuracy of the proposed method. Other PU modes are skipped because in each $2N\times 2N$ depth this PU mode is selected in $\approx 84 \%$ of the cases in the used dataset. 
To visualize this fact the histogram of PU sizes is plotted in Fig.~\ref{fig:pu} . 
This option might reduce the computational efficiency, but it can improve accuracy. 
The flowchart of the proposed algorithm is shown in Fig.~\ref{fig:flowchart}.

\section{Simulation results}
\label{sec:S_R}
In this section, the performance of proposed method is studied.
Initially, the used test set is presented, followed by the encoding setting.
Furthermore, the anchors and state-of-the-art used for comparison is also defined. Finally, the performance evaluation and complexity of the method is considered.

\subsection{Test Dataset}
Five light field lenslet images listed in Table.~\ref{tab:dataset} from EPFL Light-Field Image Dataset~\cite{EPFL} have been selected to evaluate the performance of the proposed method. 
\begin{table}[!h]
\center
\caption{Dataset selected for the evaluation.}
\label{tab:dataset}
\begin{tabular}{c|l}
   Image No. &         \multicolumn{1}{c}{Name}               \\ \hline
I01 & $Bikes$                   \\ \hline
I02 & $Danger\_de\_Mort$       \\ \hline
I04 & $Stone\_Pillars\_Outside$ \\ \hline
I09 & $Fountain\&Vincent\_2$    \\ \hline
I10 & $Friends\_1$             
\end{tabular}
\end{table}

 \begin{figure*}[!tb]
\centering
\begin{subfigure}{0.18\textwidth}
\includegraphics[width=\textwidth]{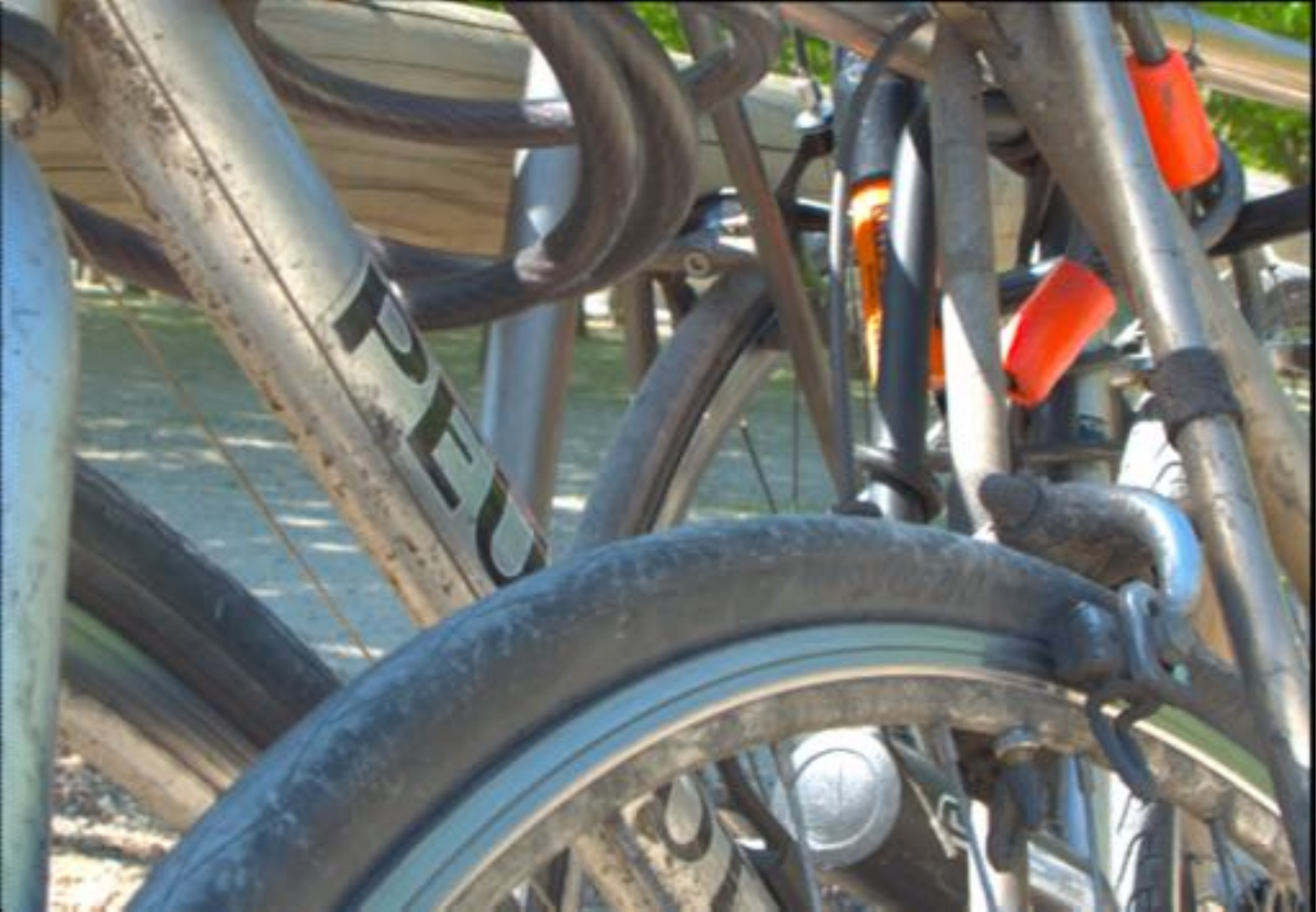}
\caption{I01}
\label{fig:c_I01}
\end{subfigure}
~ 
\begin{subfigure}{0.18\textwidth}
\includegraphics[width=\textwidth]{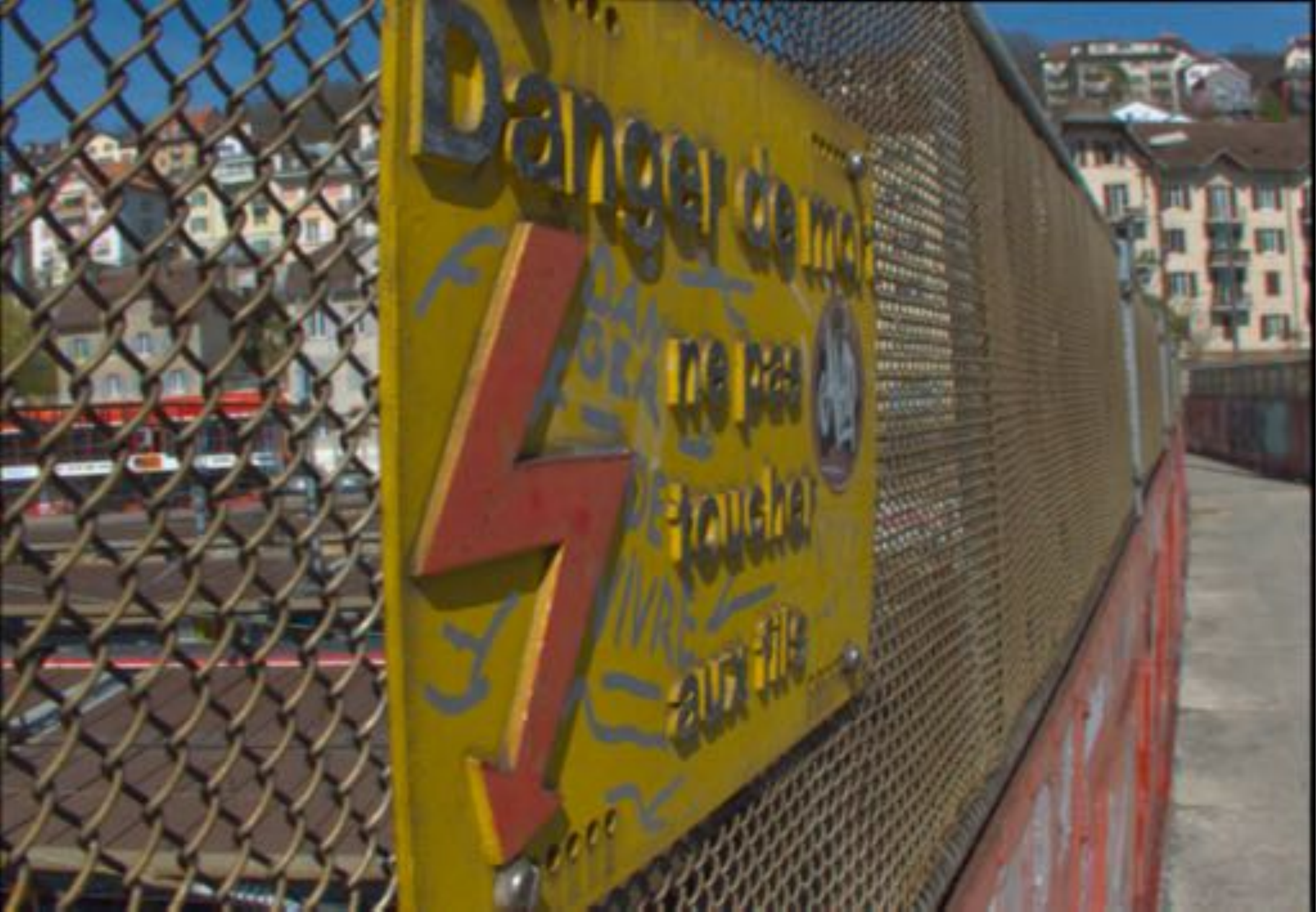}
\caption{I02}
\label{fig:c_I02}
\end{subfigure}
~ 
\begin{subfigure}{0.18\textwidth}
\includegraphics[width=\textwidth]{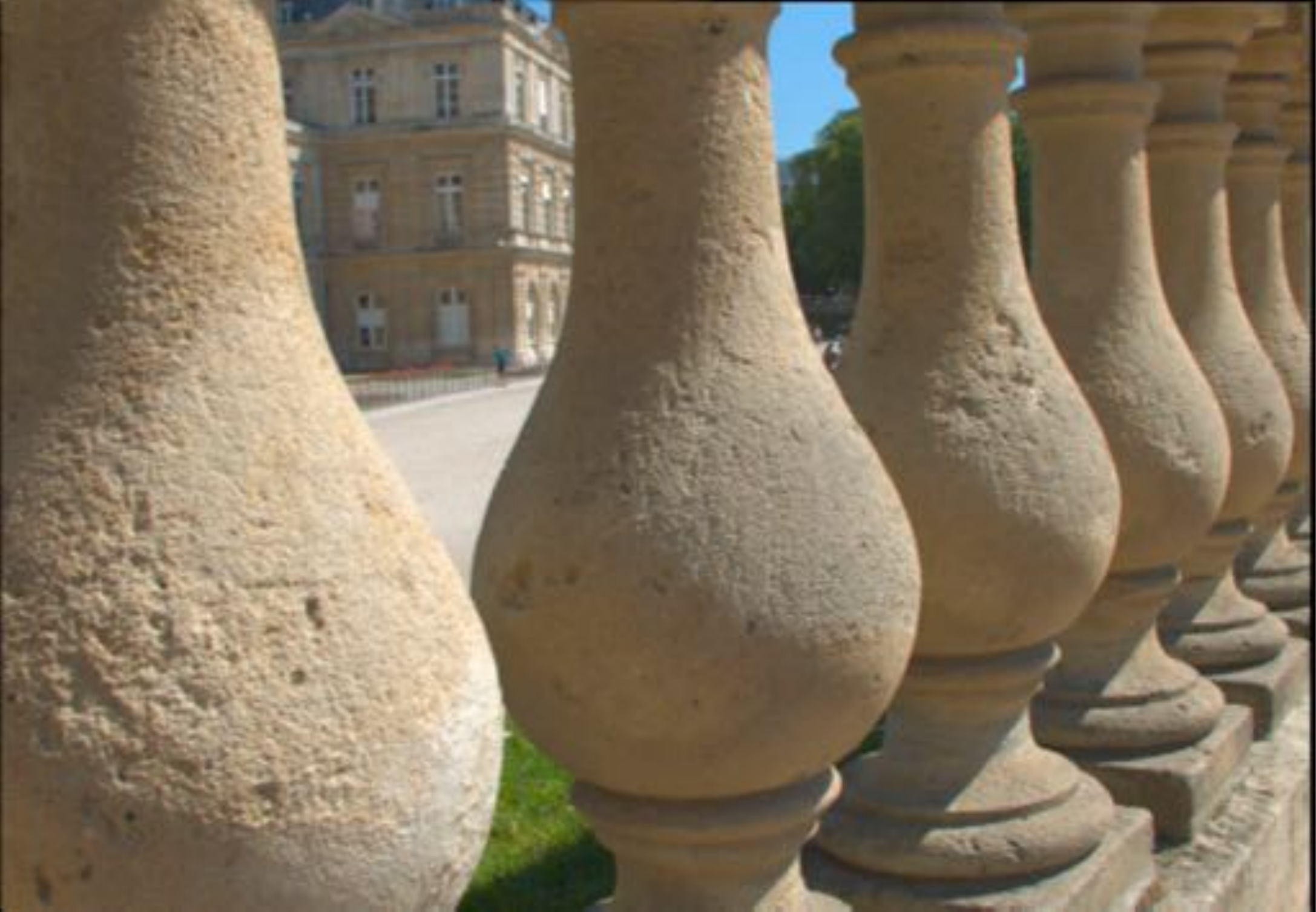}
\caption{I04}
\label{fig:c_I04}
\end{subfigure}
~ 
\begin{subfigure}{0.18\textwidth}
\includegraphics[width=\textwidth]{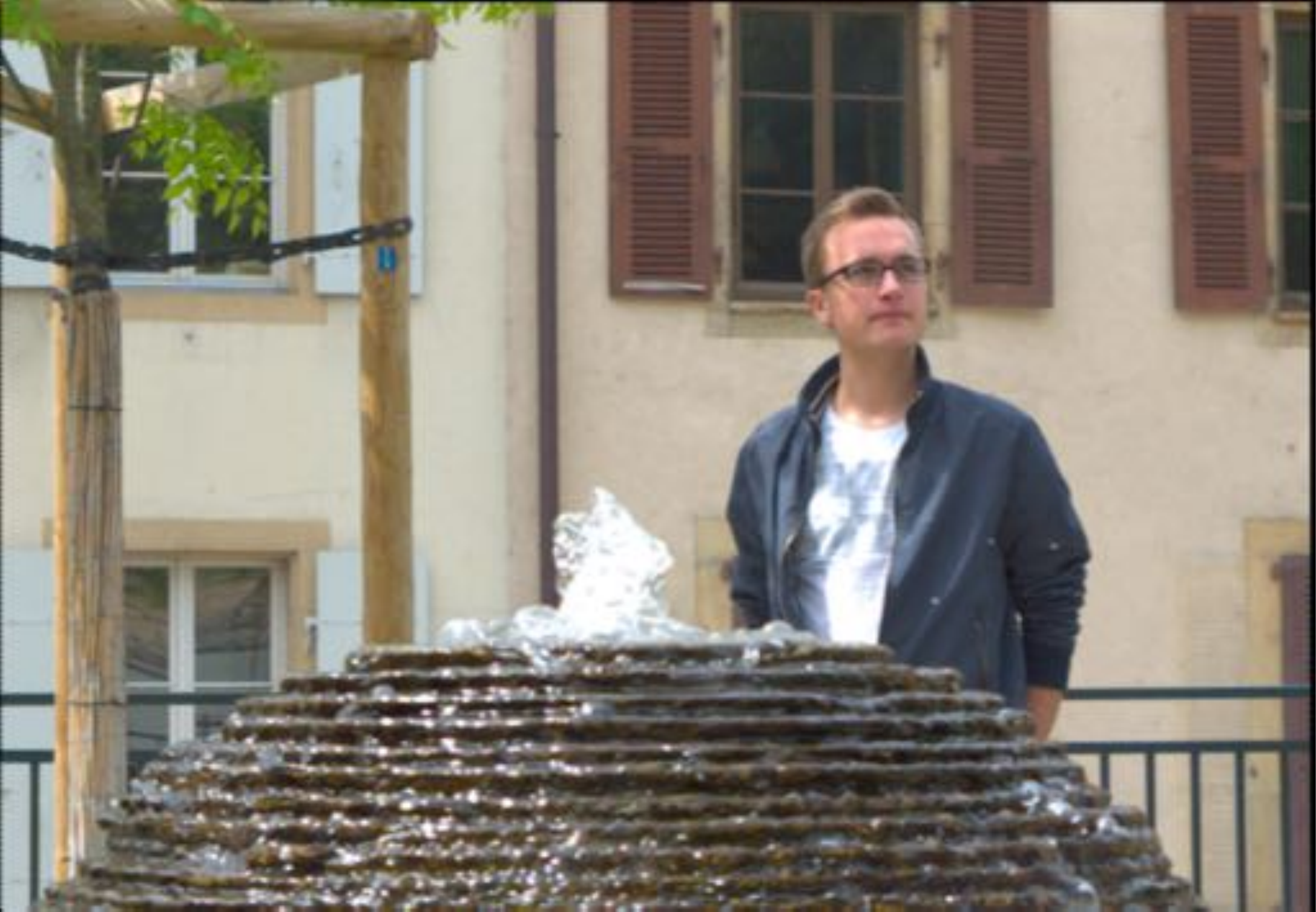}
\caption{I09}
\label{fig:c_I09}
\end{subfigure}
~ 
\begin{subfigure}{0.18\textwidth}
\includegraphics[width=\textwidth]{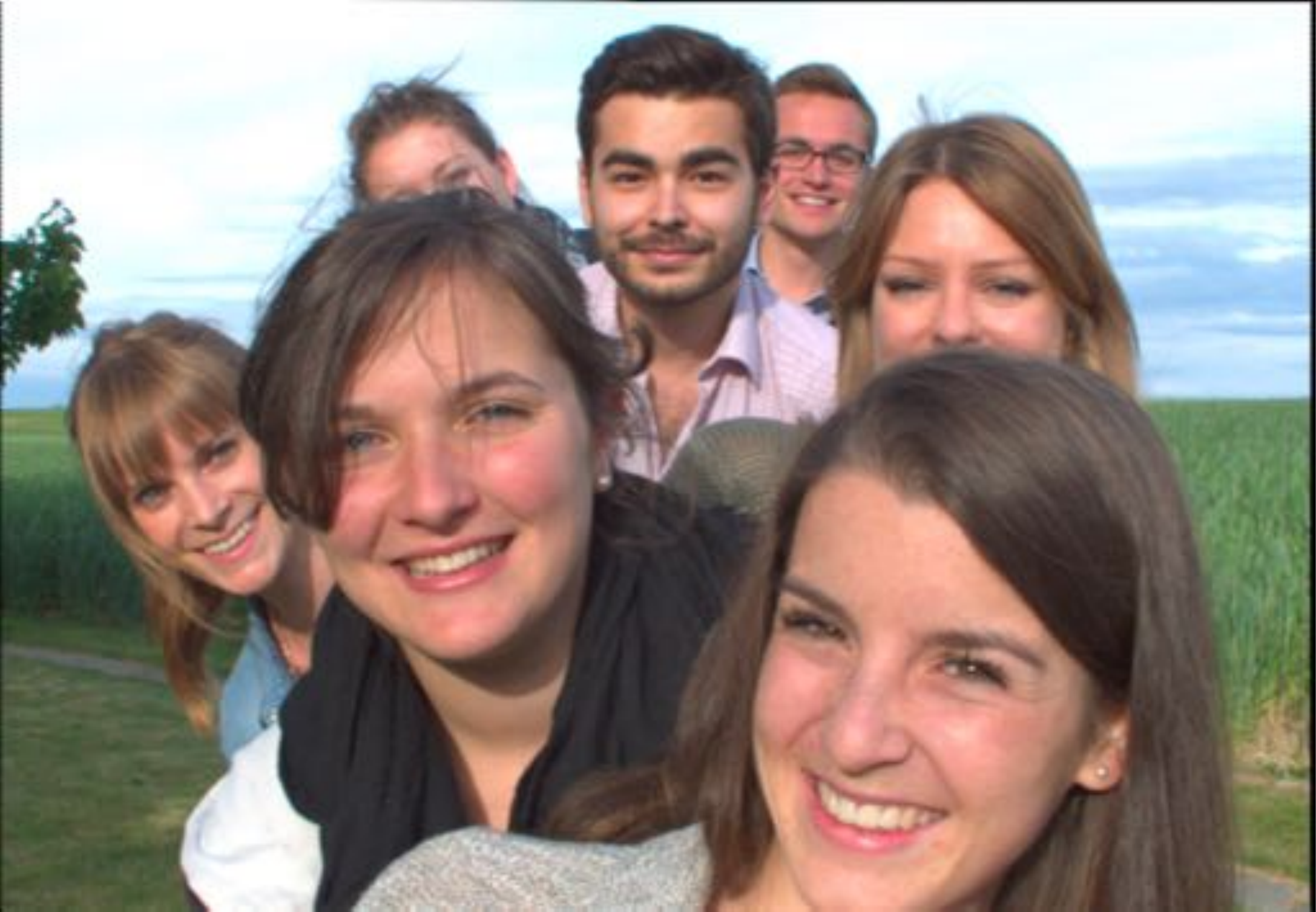}
\caption{I10}
\label{fig:c_I10}
\end{subfigure}
\caption{Central sub-aperture image views of the light fields used for testing.}
\label{fig:20}
\end{figure*}
\begin{table*}[b]
\caption{Bjontegaard relative bitrate savings (\%) and PSNR improvements (dB) over Zigzag scan order. }
\label{tab:order}
\resizebox{\textwidth}{!}{%
\begin{tabular}{c|ll|ll|ll|ll|}
    & \multicolumn{2}{c|}{Proposed} & \multicolumn{2}{c|}{Serpentine} & \multicolumn{2}{c|}{Raster} & \multicolumn{2}{c|}{Spiral} \\ \hline
    & BDRATE(\%)  & BDPSNR (dB)  & BDRATE(\%)  & BDPSNR (dB)  & BDRATE(\%)  & BDPSNR (dB)  & BDRATE(\%)  & BDPSNR (dB)  \\
I01 &      \textbf{-37.6371}       &     \textbf{1.4220}     &        -13.1765      &    0.4257      &    -12.2391         &     0.3938 &      -7.8499       &     0.2452         \\
I02 &      \textbf{-38.6638}     &       \textbf{1.4296}   &         -10.8153    &      0.3370    &       -8.4546      &       0.2610 &        -2.8901     &      0.0930    \\
I04 &      \textbf{-43.4042}      &      \textbf{1.3797}   &         -15.6835    &       0.4085   &       -14.1148      &      0.3610 &       -6.8822      &       0.1884     \\
I09 &      \textbf{-40.2363}     &       \textbf{1.4550}  &         -15.8945    &       0.4916  &       -14.3282      &     0.4418 &         -8.0453    &        0.2497      \\
I10 &      \textbf{-38.7987}      &      \textbf{1.2766}  &         -11.0981    &       0.3033  &       -10.0967      &    0.2739 &       -5.3532      &        0.1488    \\
Average& \textbf{-39.7480} &    \textbf{1.3926}  &               -13.3336  &       0.3932  &        -11.8467     &  0.3463  &           -6.2041  &        0.1850  
\end{tabular}}
\end{table*}

The central sub-aperture image views extracted from each lenslet image are shown in Fig.~\ref{fig:20}. These images contain different textures, disparities and amount of redundancy within sub-aperture images which are measured by the Geometric Space-View Redundancies (GSVR) descriptor~\cite{GSVR}. This ensures that the selected dataset can cover different types of light field images ranging from low to high redundancy.

\subsection{Anchor and State-of-the-art}

The proposed method is initially compared with the conventional scan orders using zigzag as an anchor. The same QP values are used for all the sub-aperture images in this comparison to show the effect of the scanning order.

To assess the performance of the proposed method, the lenslet pseudo-sequences are arranged in a serpentine scanning order were also compressed with HEVC/x.265 and VP9~\cite{vp9}. The serpentine order was selected because it gives a better performance than other scan orders. Moreover, three state-of-the-art methods, Linear Approximation Prior ($LAP$)~\cite{lap}, MuLE~\cite{mule, vm}, and WaSP~\cite{wasp,vm} are also selected to compare with the proposed method.

Linear Approximation Prior ($LAP$)~\cite{lap} exploits linearity between the sub-aperture images. In LAP, a subset of views are ordered as a pseudo-sequence and are compressed using HEVC. Then, they are decoded and are used to estimate the remaining views. Warping and Sparse Prediction ($WaSP$)~\cite{wasp}, which is also known as JPEG Pleno  Verification Model  1 (VM1), uses hierarchical warping, merging, and sparse prediction to encode the light fields.
MUltidimensional Lightfield Encoder ($MuLe$), which is used in JPEG Pleno Verification Model 2 (VM2), exploits the 4D redundancy of light fields by using a 4D transform and hexadeca-trees. In MuLe, the light field is divided into 4D blocks and a 4D Discrete Cosine Transform is computed. Then, hexadeca-trees on a bitplane-by-bitplane basis are used to generate transform coefficients of the 4D block, and an adaptive arithmetic coder is used to encode the bitstream.

\subsection{Coding Condition}
The light field toolbox v0.4~\cite{toolbox} of Matlab is used to decompose raw lenslet images into $15\times15$ sub-aperture images of $625\times434$ pixels resolution and 10 bits depth per color channel. 
Sub-aperture views are padded to $626\times434$ pixels resolution and are converted to YCbCr format using ITU-R Recommendation BT.709-6~\cite{itu709}. The proposed method has been implemented with no chromatic subsampling 4:4:4 and with 4:2:2 chromatic subsampling. 
The case without any subsampling (4:4:4) is compared with the $MuLe$ results because the used implementation does not allow any chromatic subsampling. All the other cases used 4:2:2 chromatic sub-sampling.

HEVC HM reference software version 16.18\footnote{https://hevc.hhi.fraunhofer.de} has been used to implement the proposed method.  
Four bitrates targeting  $R1=0.75~bpp$, $R2=0.1~bpp$, $R3=0.02~bpp$ and $R4=0.005~bpp$ have been selected. 

\begin{figure*}[!t]
\begin{minipage}[t]{0.19\linewidth}
\centering
\centerline{\includegraphics[width=\textwidth,height=0.7\textwidth,trim={3cm 8cm 4cm 9cm},clip]{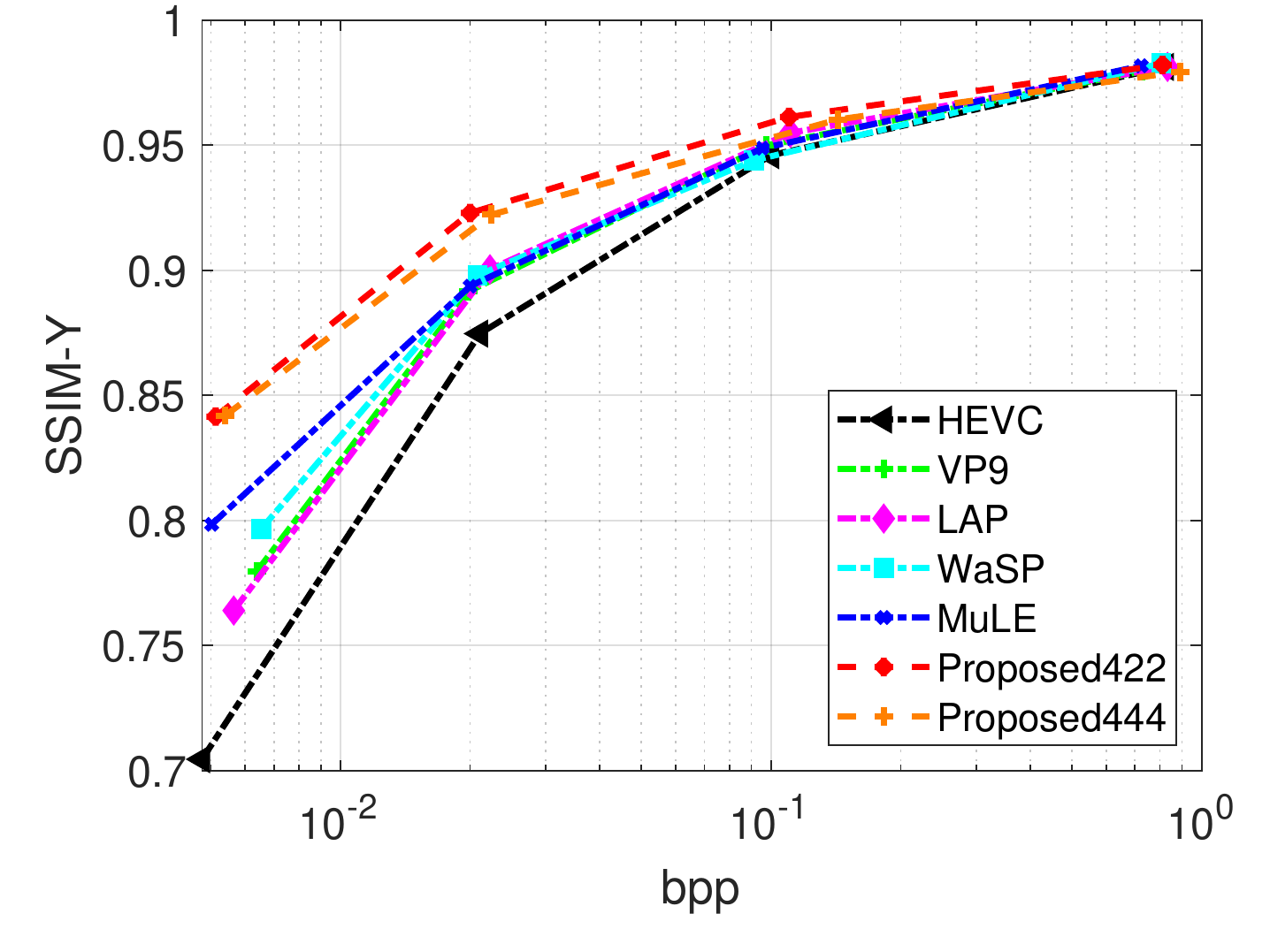}}
\subcaption{I01}
\end{minipage}
\begin{minipage}[t]{0.19\linewidth}
\centering
\centerline{\includegraphics[width=\textwidth,height=0.7\textwidth,trim={3cm 8cm 4cm 9cm},clip]{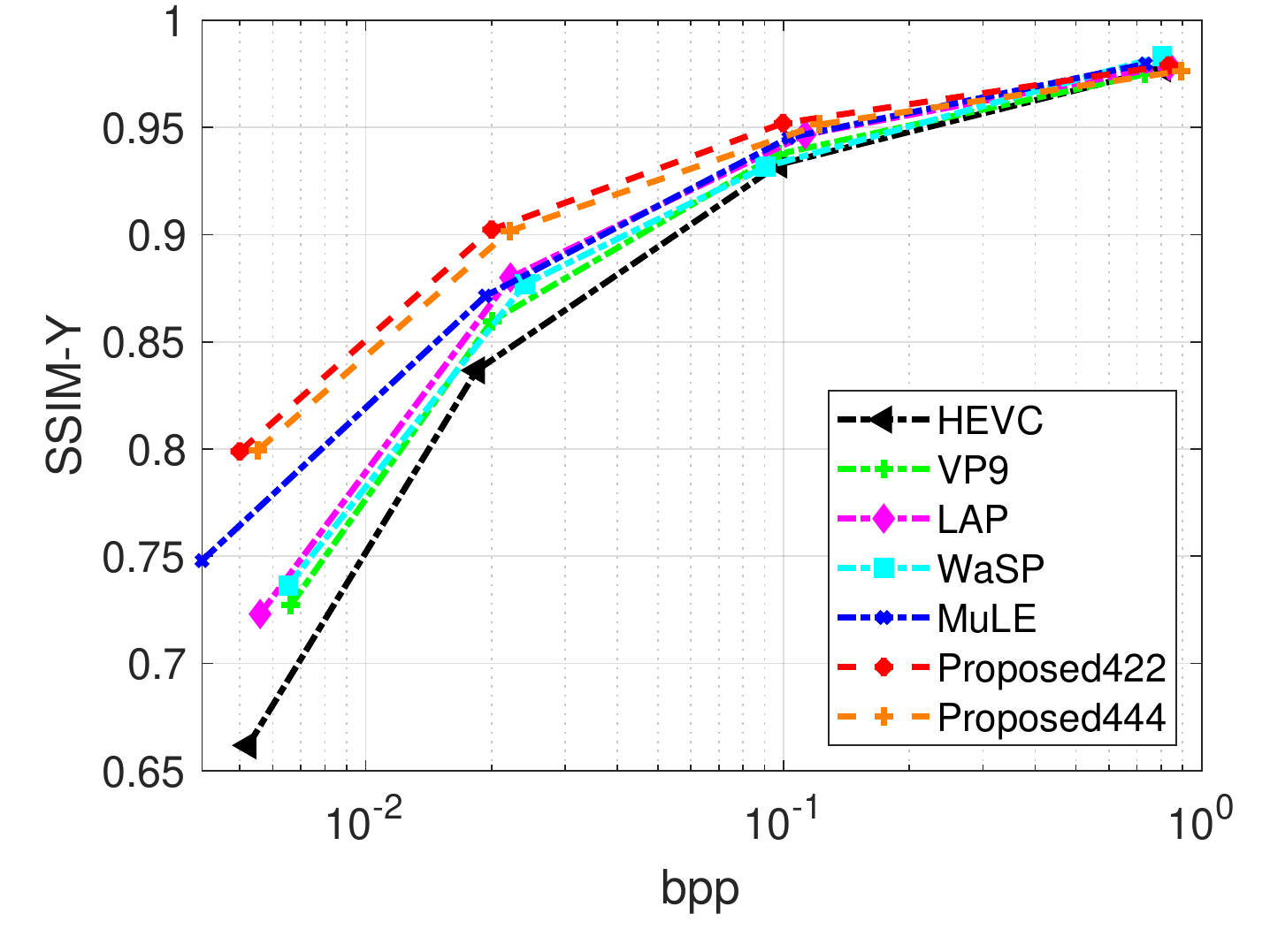}}
\subcaption{I02}
\end{minipage}
\begin{minipage}[t]{0.19\linewidth}
\centering
\centerline{\includegraphics[width=\textwidth,height=0.7\textwidth,trim={3cm 8cm 4cm 9cm},clip]{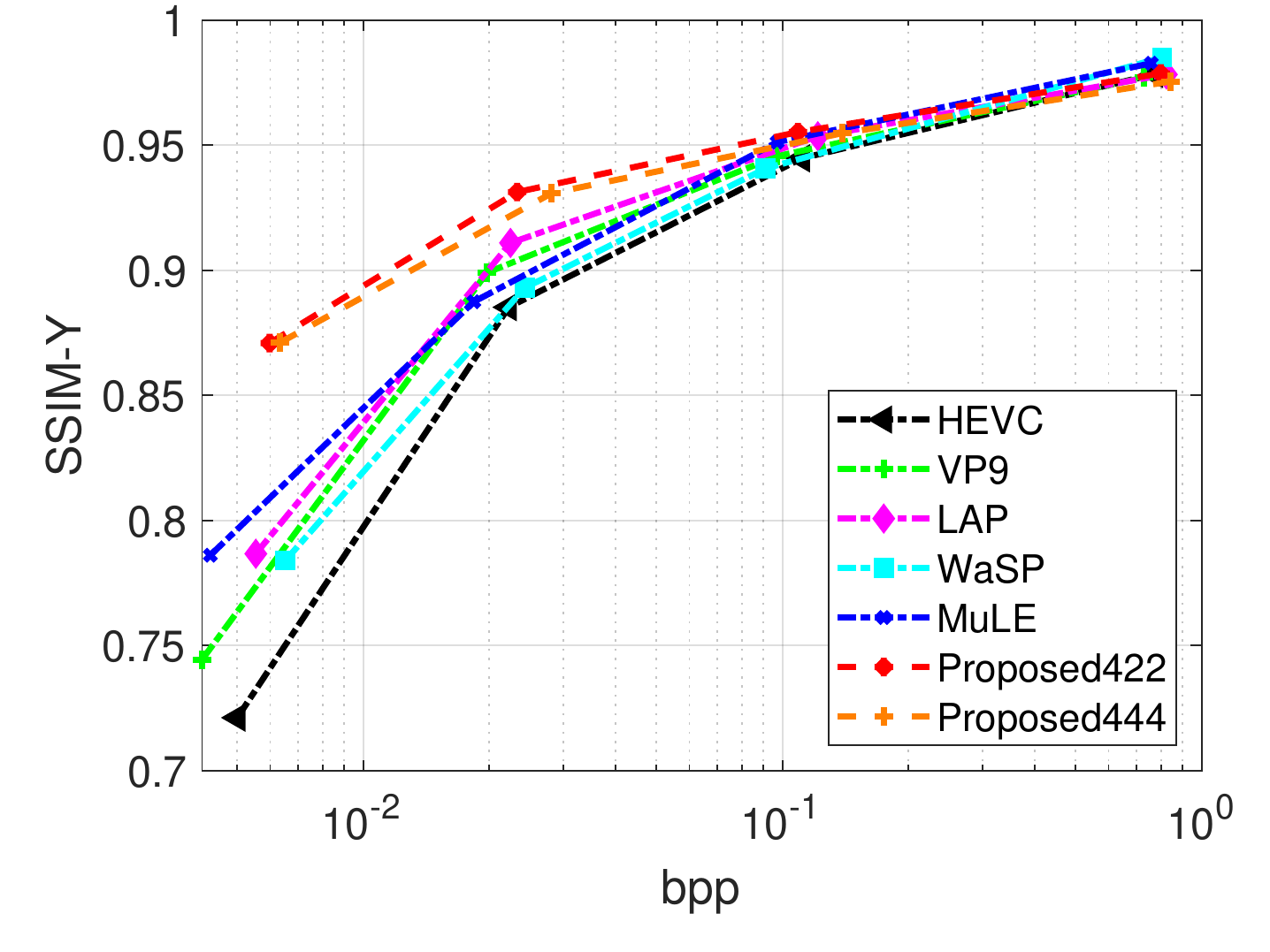}}
\subcaption{I04}
\end{minipage}
\begin{minipage}[t]{0.19\linewidth}
\centering
\centerline{\includegraphics[width=\textwidth,height=0.7\textwidth,trim={3cm 8cm 4cm 9cm},clip]{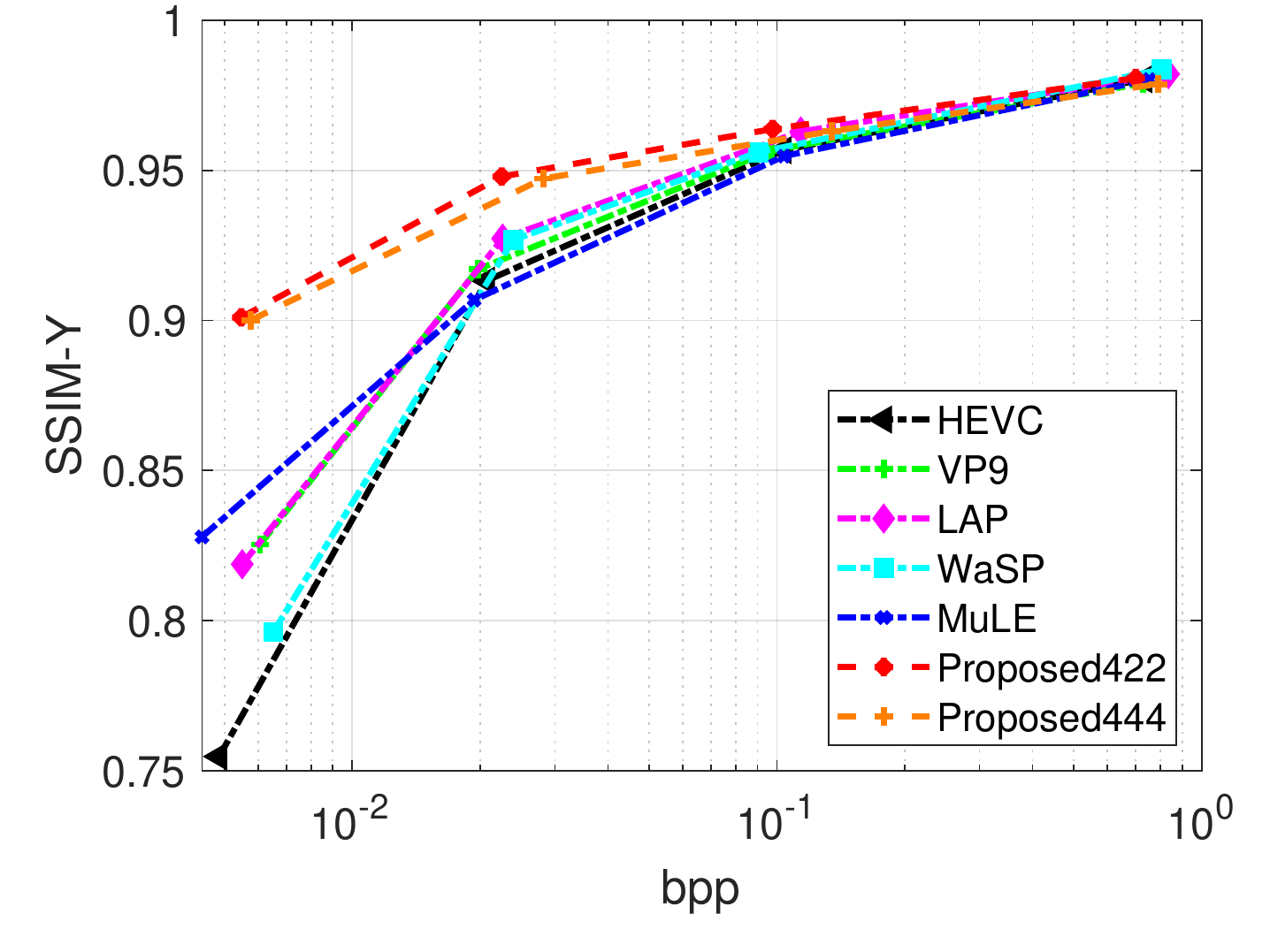}}
\subcaption{I09}
\end{minipage}
\begin{minipage}[t]{0.19\linewidth}
\centering
\centerline{\includegraphics[width=\textwidth,height=0.7\textwidth,trim={3cm 8cm 4cm 9cm},clip]{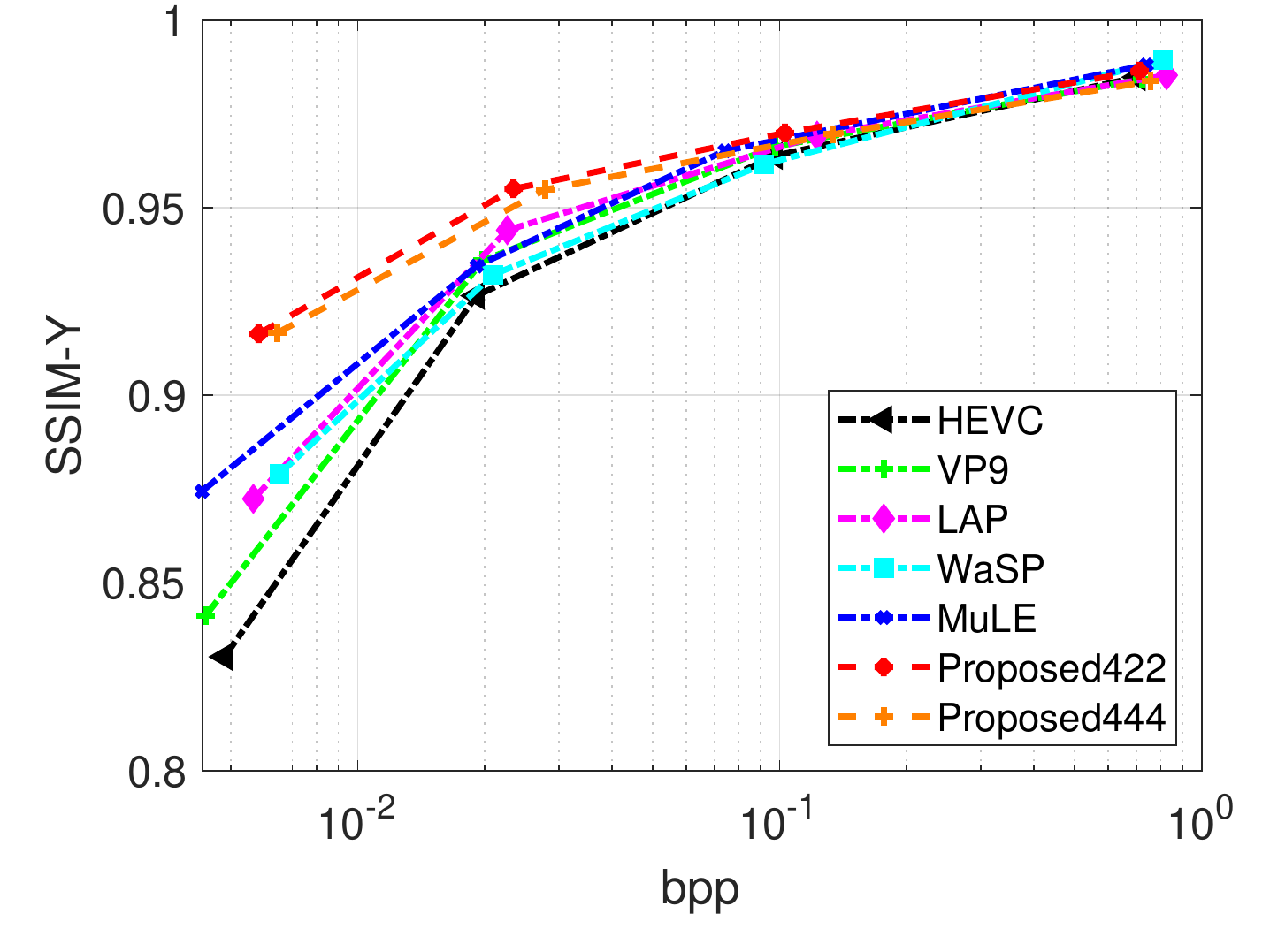}}
\subcaption{I10}
\end{minipage}
\caption{Rate-distortion curves (SSIM-Y vs. bpp).}
\label{fig:rds}

\end{figure*}
\begin{figure*}[!t]
\begin{minipage}[t]{0.19\linewidth}
\centering
\centerline{\includegraphics[width=\textwidth,height=0.7\textwidth,trim={3.8cm 8cm 4.3cm 9cm},clip]{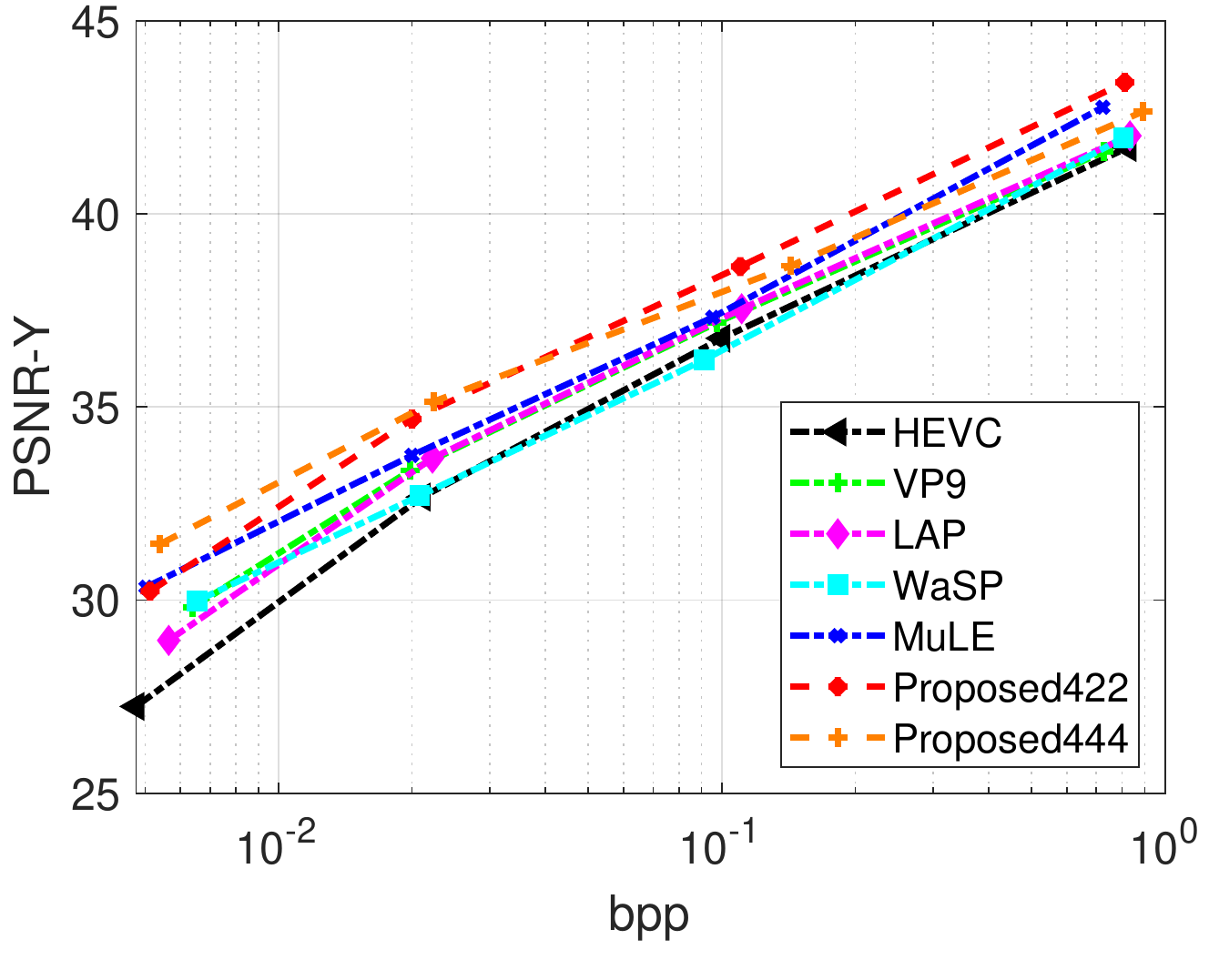}}
\subcaption{I01}
\end{minipage}
\begin{minipage}[t]{0.19\linewidth}
\centering
\centerline{\includegraphics[width=\textwidth,height=0.7\textwidth,trim={3.8cm 8cm 4.3cm 9cm},clip]{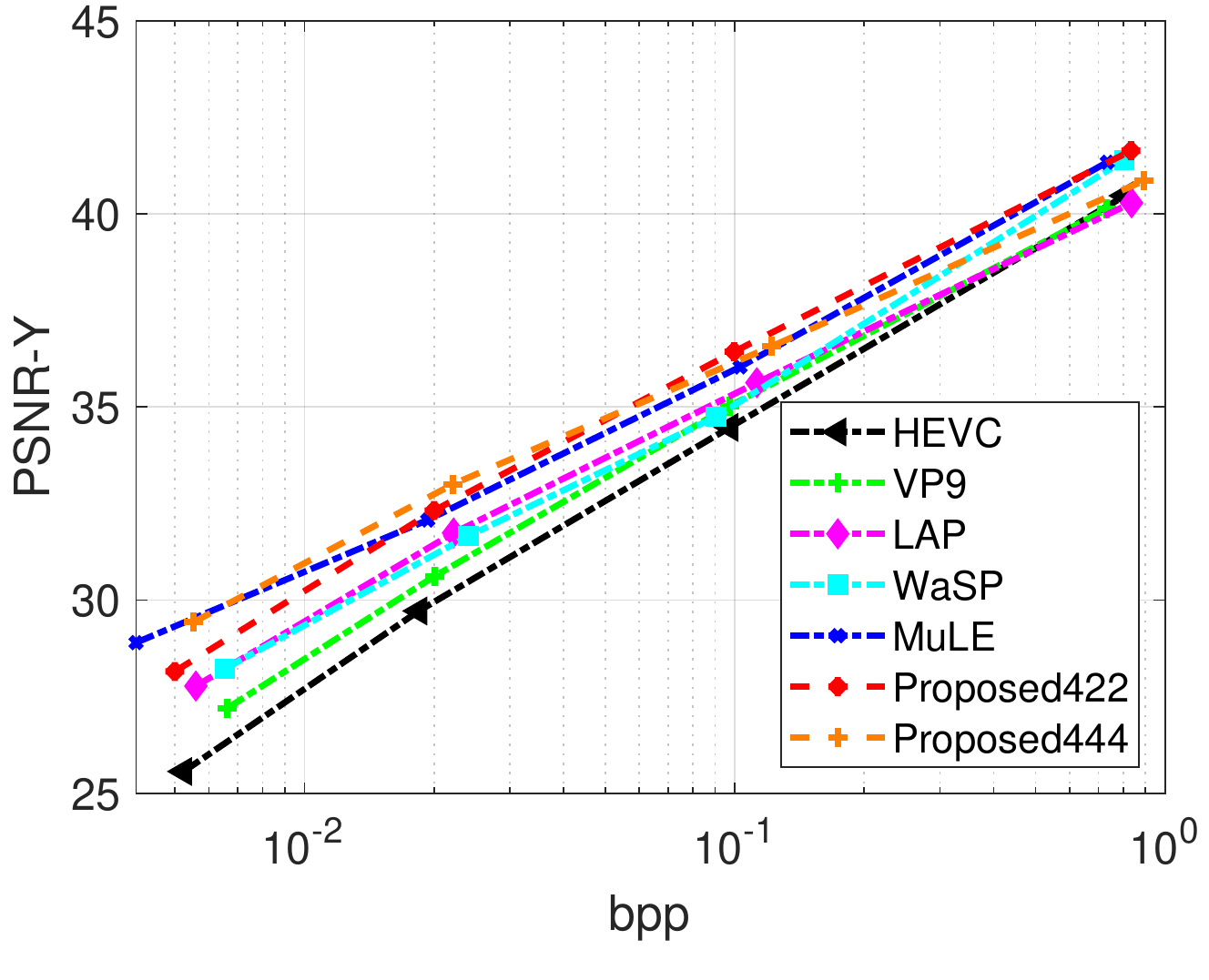}}
\subcaption{I02}
\end{minipage}
\begin{minipage}[t]{0.19\linewidth}
\centering
\centerline{\includegraphics[width=\textwidth,height=0.7\textwidth,trim={3.8cm 8cm 4.3cm 9cm},clip]{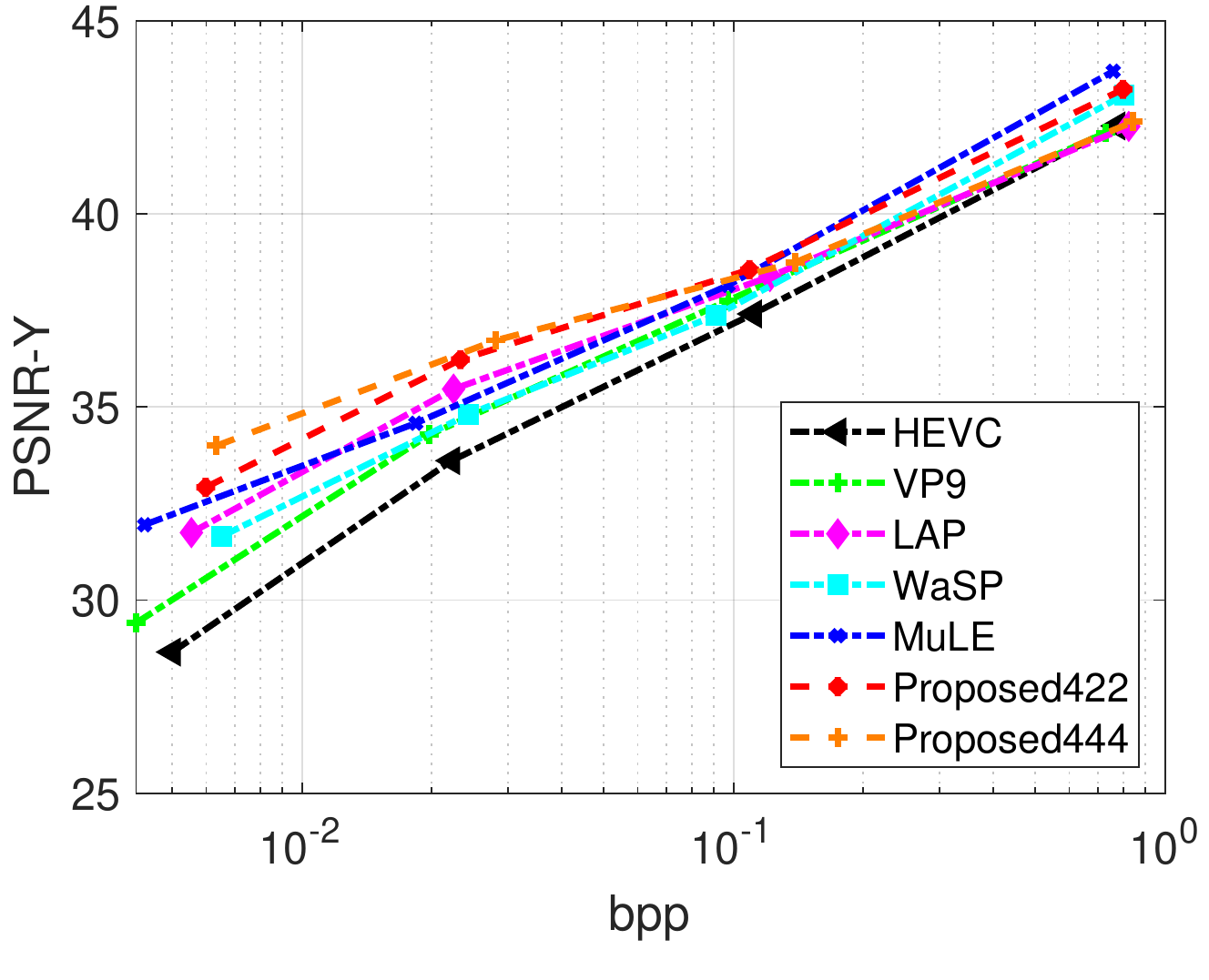}}
\subcaption{I04}
\end{minipage}
\begin{minipage}[t]{0.19\linewidth}
\centering
\centerline{\includegraphics[width=\textwidth,height=0.7\textwidth,trim={3.8cm 8cm 4.3cm 9cm},clip]{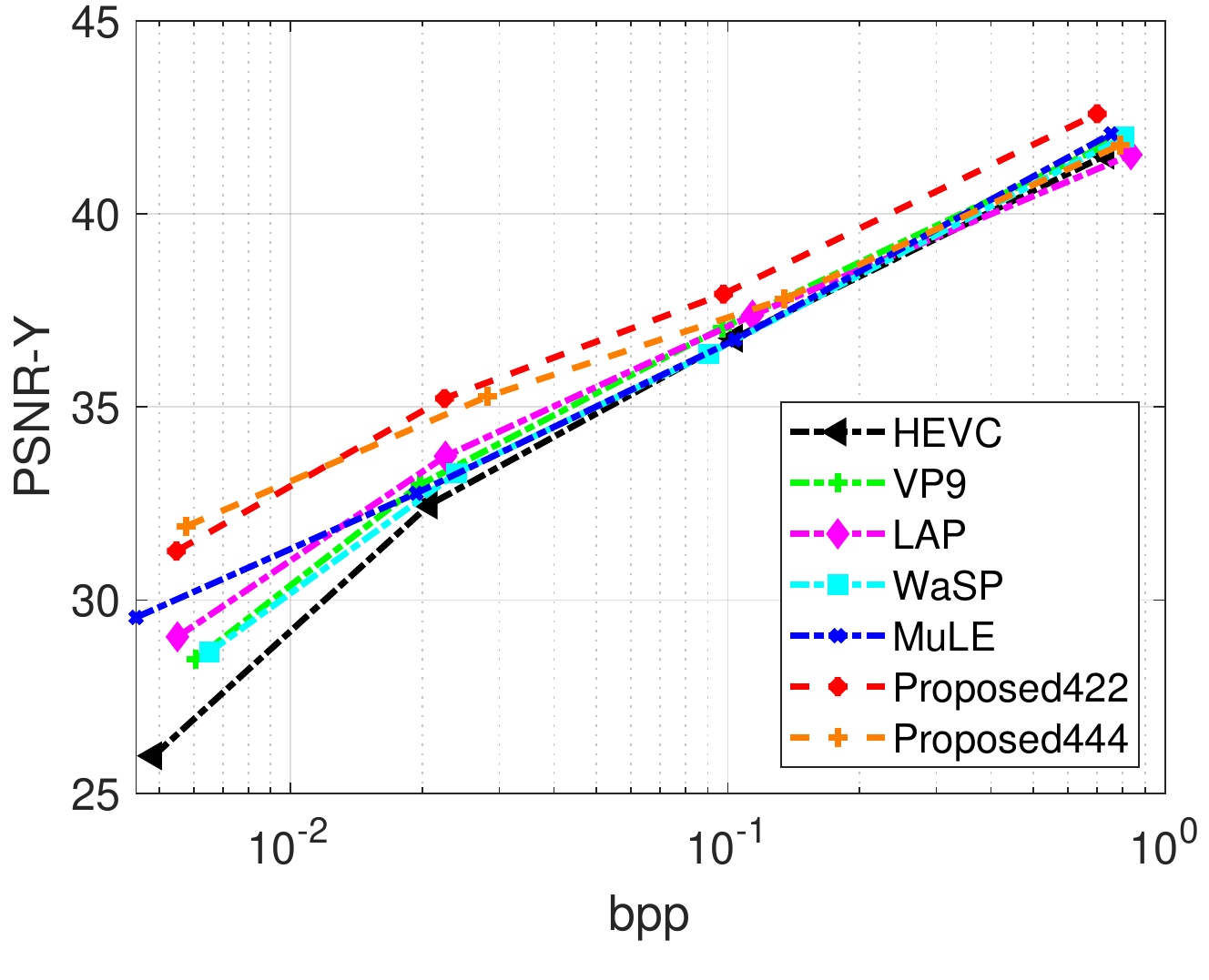}}
\subcaption{I09}
\end{minipage}
\begin{minipage}[t]{0.19\linewidth}
\centering
\centerline{\includegraphics[width=\textwidth,height=0.7\textwidth,trim={3.8cm 8cm 4.3cm 9cm},clip]{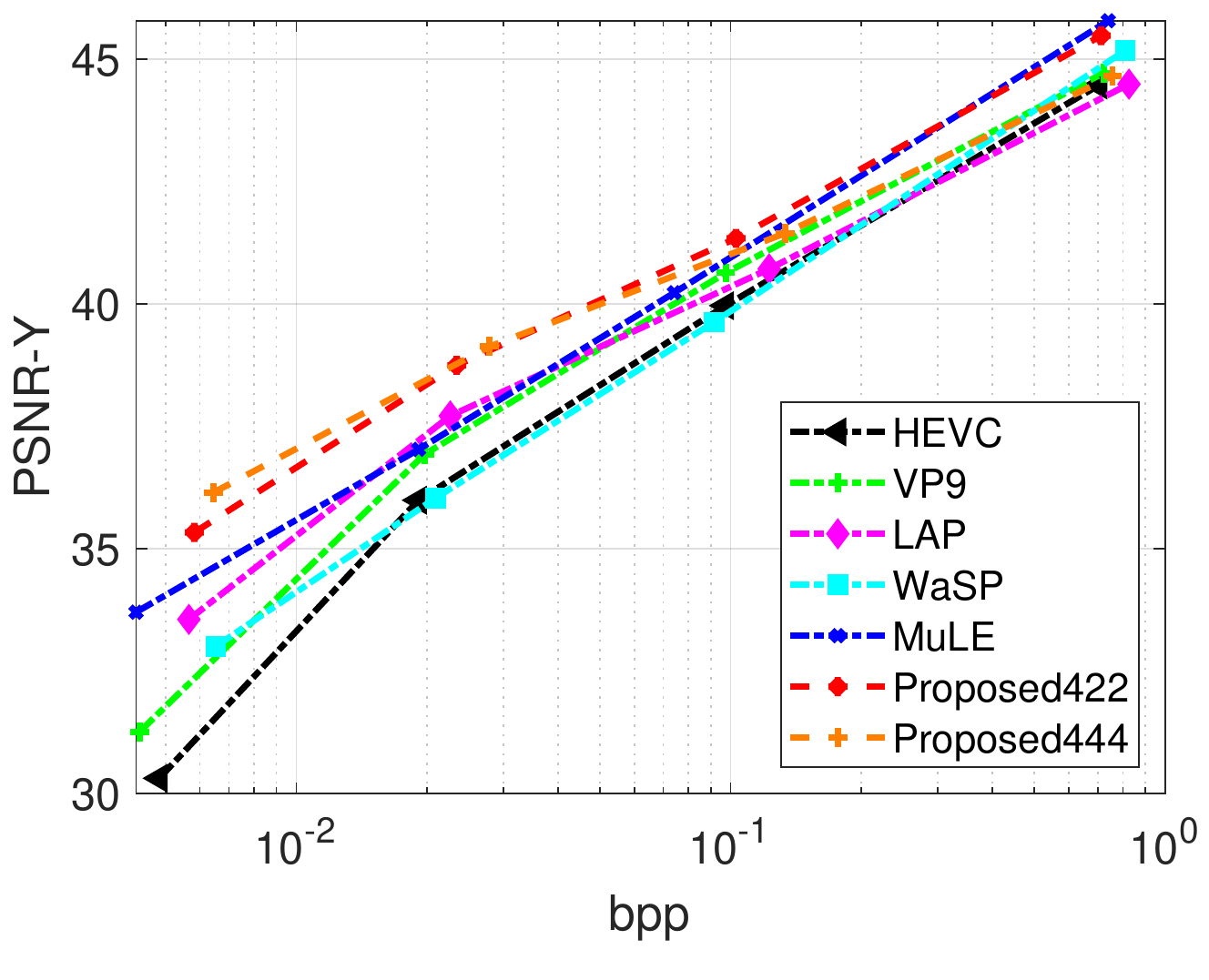}}
\subcaption{I10}
\end{minipage}
\caption{Rate-distortion curves (PSNR-Y vs. bpp).}
\label{fig:rdp}

\end{figure*}
%


\subsection{Coding efficiency }
Table~\ref{tab:order} summarizes RD performance of the different scan orders against the zigzag method in terms of BDBR (\%) and BDPSNR (dB)~\cite{bjonteegard} using the metric $PSNR_Y$ and 4:2:2 chromatic subsampling.
The compression of the most central view, which is intra coded, plays an important role in the performance of the proposed method. A lower QP is given to the central image to improve the proposed method compression gain.

Figs.~\ref{fig:rds} and~\ref{fig:rdp} show rate-distortion curves for all the contents using $SSIM_Y$ and $PSNR_Y$, respectively.
From the observation of the different plots it becomes obvious that the compression efficiency of the proposed method largely surpasses the serpentine scan order. Compression efficiency is also higher than $LAP$, $Mule$, and $WasP$.

\subsection{Complexity }
The proposed method has been implemented in three ways: \\
1-  Conventional serial computing as anchor without implementing the proposed depth decision method. This is considered as a representation of the pseudo-sequence methods. These methods cannot use parallel computing at a frame level due to the dependency among the sub-aperture images. 
\\
\noindent 2- Using the proposed CTU partitioning method for depth decision with conventional serial computing to show the time-complexity reduction ($\Delta T_s$).
\\
\noindent  3- Using the proposed CTU partitioning method for depth decision and encoding each region independently in a parallel way ($\Delta T_p$).

\begin{figure}[!h]
    \centering
    \begin{subfigure}[b]{0.43\textwidth}
        \includegraphics[width=\textwidth]{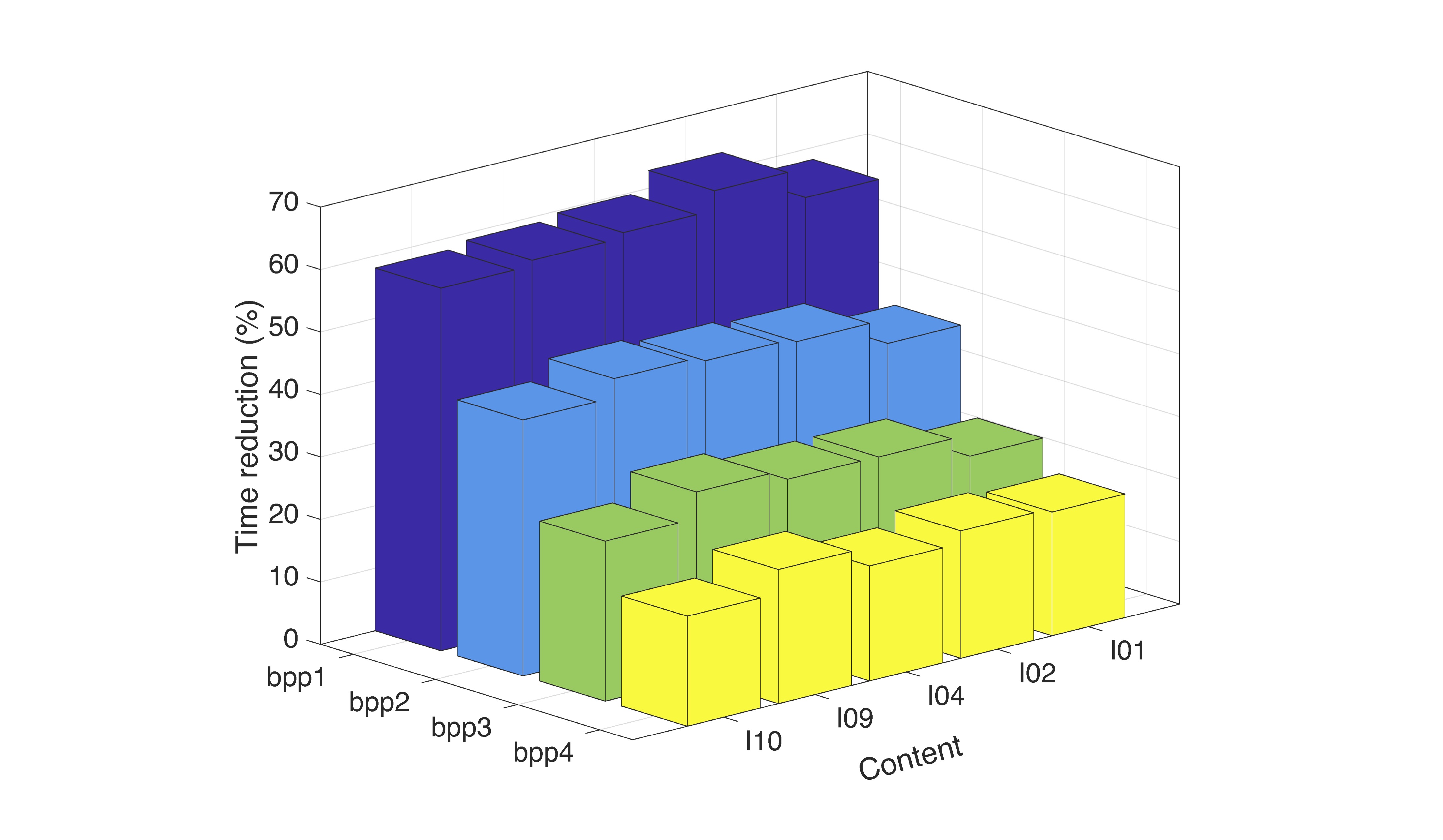}
        \caption{Serial computing}
        \label{fig:serial}
    \end{subfigure}
    ~ 
    \begin{subfigure}[b]{0.42\textwidth}
        \includegraphics[width=\textwidth]{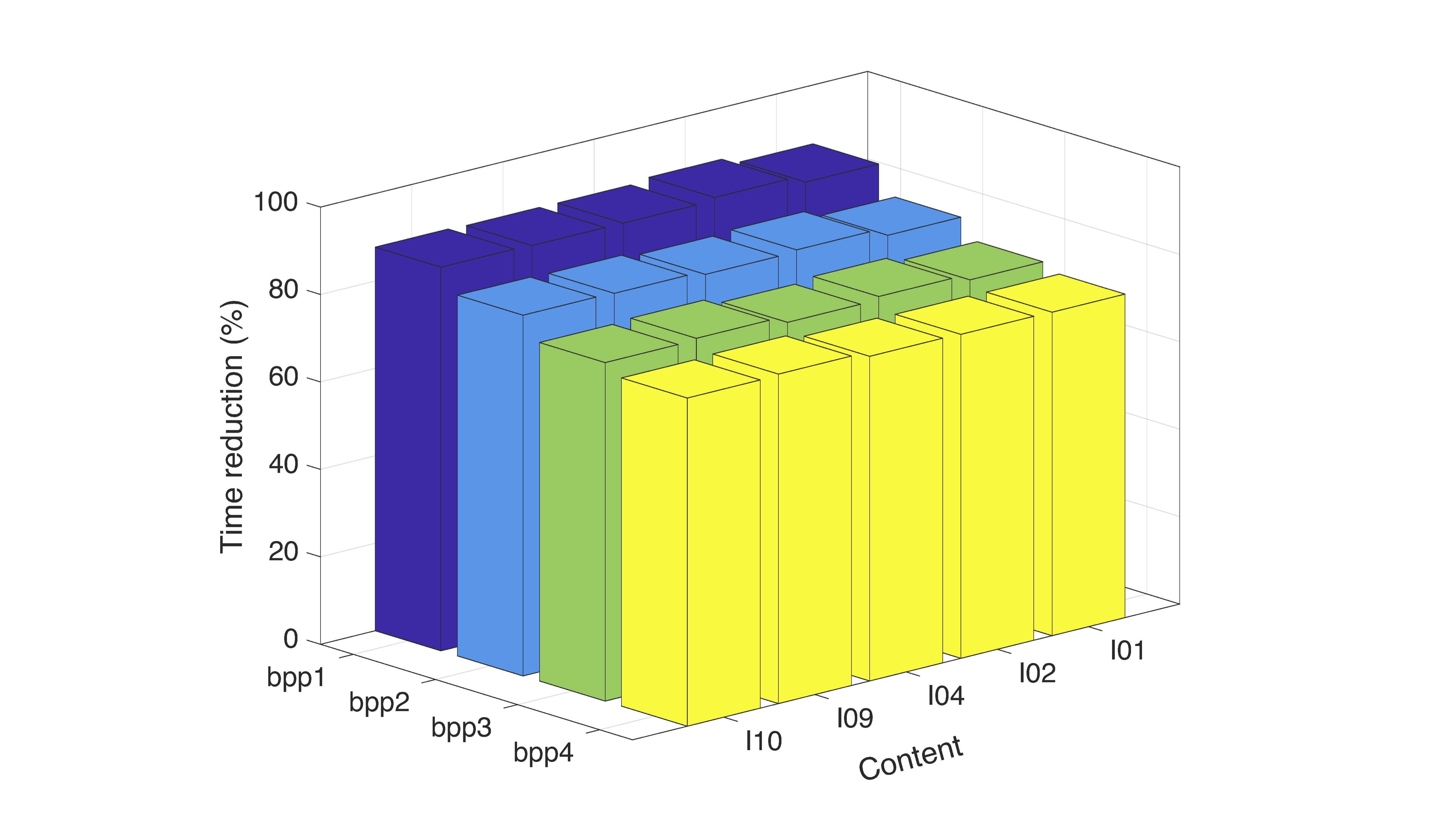}
        \caption{Parallel computing}
        \label{fig:parallel}
    \end{subfigure}
    \caption{Time reduction of the proposed method using~(\subref{fig:serial}) serial and~(\subref{fig:parallel}) parallel computing. }
    \label{fig:time}
\end{figure}

Time reduction for cases 2 and 3 at different bitrates and images are plotted in Fig.~\ref{fig:time}.

Experimental results are given in Table~\ref{tab:4}. The results indicate that the proposed method can reduce the time complexity of the pseudo-sequence based methods while the quality loss is negligible.
As can be seen in Table 1, the proposed CTU partitioning method for depth decision reduces the computational complexity by around 37\%. Adding parallel computing will further reduce the time complexity to levels near 80\% on average.
\begin{table}[!h]
\caption{Comparison of encoding time and coding efficiency}
\label{tab:4}
\center
\begin{tabular}{c|ccccl}
    & $BDPSNR (dB)$ & $BDBR (\%)$ & $\Delta T_p (\%)$ & $\Delta T_s (\%)$ \\ \hline
I01 &  -0.0238       & 0.9358      & 79.0629    &   35.3418  \\ \hline
I02 & -0.0092       & 0.3379      & 80.2999 &  38.5218       \\ \hline
I04 & -0.0056       & 0.2646     & 79.8682 &   37.3928     \\ \hline
I09 & -0.0123       & 0.5467      & 80.7433      &  38.5478 \\ \hline
I10 & -0.0150       & 0.6624      & 80.6744    &   35.5681  \\ \hline
Average &-0.0132      &0.5495      & 80.1297       & 37.0744 \\ \hline                                              
\end{tabular}
\end{table}

\section{Conclusion}
\label{sec:conclusion}
In this paper, a lenslet image compression method based on the HEVC codec was proposed. The method uses the decomposition of the lenslet image into multi sub-aperture images and considers them as sub-aperture pseudo-sequences to be compressed by an HEVC video codec. To decrease the average distances of current sub-aperture images and their references, the sub-aperture images were divided into four regions. Each region is compressed independently using the central sub-aperture image as the first reference image. 
To exploit the similarity among the adjacent sub-aperture images a new GOP structure was introduced. 
It selects sub-aperture images based on their distances as active references for prediction. Due to higher similarity among the spatially closer references, their co-located CTUs are used to predict depth of the current CTU.
Simulation results show that the proposed method outperforms conventional methods in terms of rate-distortion, as well as in terms of complexity.
Independent encoding of each region enables parallel processing and results in larger computational efficiency.

\bibliographystyle{IEEEtran}
\bibliography{refs}

\end{document}